\documentclass[journal]{IEEEtran}

\usepackage{epsfig}
\usepackage{graphicx}
\usepackage{amsmath}
\usepackage{amssymb}
\usepackage{dsfont}
\usepackage{gensymb}
\usepackage{comment}
\usepackage{subfigure}
\usepackage{pifont}
\usepackage{multirow}
\usepackage[multiple]{footmisc}
\usepackage{tabularx,ragged2e,booktabs}
\usepackage{pbox}
\usepackage[ruled,vlined]{algorithm2e}
\usepackage{algpseudocode,url}

\def\etal{{et. al }}
\def\tabletext{\small}
\def\panoh{{Panoramic Hyperlapse }}
\def\panohns{{Panoramic Hyperlapse}}
\def\egos{{EgoSampling }}
\def\egosns{{EgoSampling}}
\def\naive{{na\"{\i}ve} }

\newcommand{\specialcell}[2][c]{%
  \begin{tabular}[#1]{@{}c@{}}#2\end{tabular}}

\begin{document}
\title{\egosns: Wide View Hyperlapse from Egocentric Videos}
\author{Tavi~Halperin,  Yair~Poleg,   Chetan~Arora, and Shmuel~Peleg
\thanks{This research was supported by Israel Ministry of Science, by Israel Science Foundation, by DFG, by Intel ICRI-CI, and by Google.}
\thanks{Tavi Halperin, Yair Poleg, and Shmuel Peleg are with The Hebrew University of Jerusalem, Israel.}
\thanks{Chetan Arora is with IIIT Delhi, India.}
}

\maketitle

\begin{abstract}
    The possibility of sharing one's point of view makes use of wearable cameras compelling. These videos are often long, boring and coupled with extreme shake, as the camera is worn on a moving person. Fast forwarding (i.e. frame sampling) is a natural choice for quick video browsing. However, this accentuates the shake caused by natural head motion in an egocentric video, making the fast forwarded video useless. We propose \egosns, an adaptive frame sampling that gives stable, fast forwarded, hyperlapse videos. Adaptive frame sampling is formulated as an energy minimization problem, whose optimal solution can be found in polynomial time. We further turn the camera shake from a drawback into a feature, enabling the increase in field-of-view of the output video. This is obtained when each output frame is mosaiced from several input frames. The proposed technique also enables the generation of a single hyperlapse video from multiple egocentric videos, allowing even faster video consumption.
\end{abstract}


\section{Introduction}\label{sec:introduction}

\IEEEPARstart{W}{hile} the use of egocentric cameras is on the rise, watching raw egocentric videos is unpleasant. These videos, captured in an `always-on' mode, tend to be long, boring, and unstable. Video summarization \cite{grauman-story, grauman-important-people, rehg_gaze}, temporal segmentation \cite{us_egoseg, compact_cnn} and action recognition \cite{kitani, ryoo_pooled} methods can help browse and consume large amount of egocentric videos. However, these algorithms  make strong assumptions in order to work properly (e.g. faces are more important than unidentified blurred images). The information produced by these algorithms helps the user skip most of the input video. Yet, the only way to watch a video from start to end, without making strong assumptions, is to play it in a fast-forward manner. However, the natural camera shake gets amplified in \naive fast-forward (i.e. frame sampling). An exceptional tool for generating stable fast forward video is the recently proposed ``Hyperlapse" \cite{hyperlapse}. Our work was inspired by \cite{hyperlapse}, but take a different, lighter, approach.

\begin{figure}[t]
    \centering
    \subfigure[]{\includegraphics[width=0.9\columnwidth]{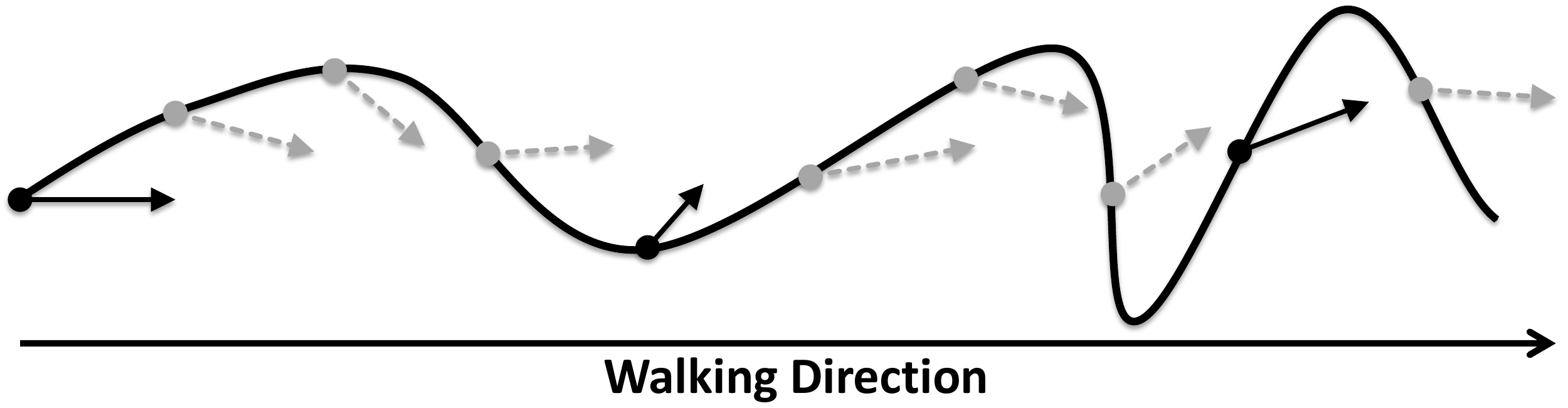}} \\
    \subfigure[]{\includegraphics[width=0.9\columnwidth]{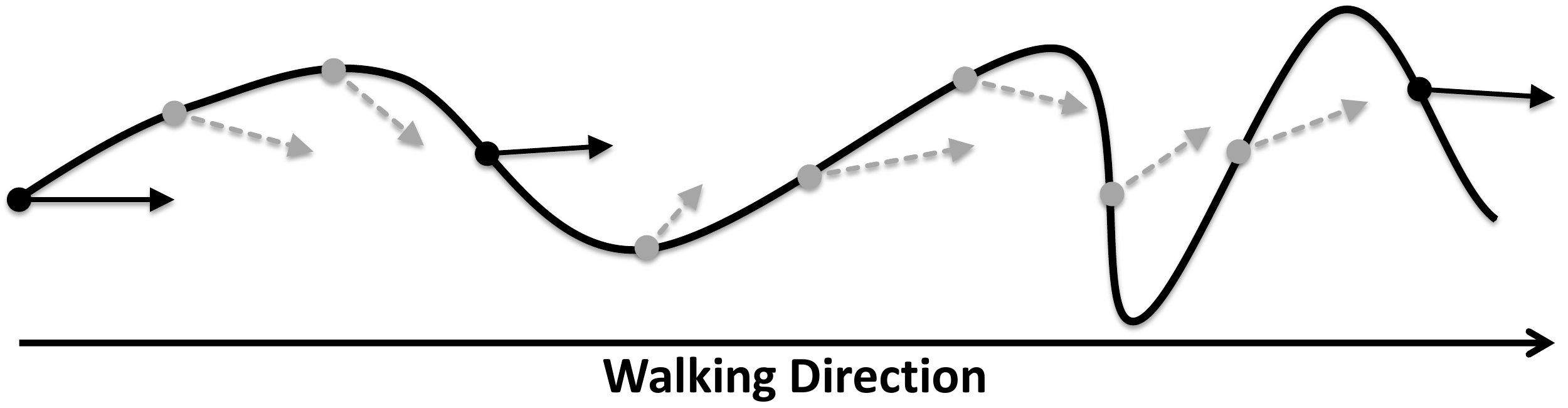}}
    \caption{Frame sampling for Fast Forward. A view from above on the camera path (the line) and the viewing directions of the frames (the arrows) as the camera wearer walks forward during a couple of seconds. (a) Uniform $5\times$ frames sampling, shown with solid arrows, gives output with significant changes in viewing directions. (b) Our frame sampling, represented as solid arrows, prefers forward looking frames at the cost of somewhat non uniform sampling.}
    \label{fig:ff-schematic}
\end{figure}

Fast forward is a natural choice for faster browsing of videos. While \naive fast forward uses uniform frame sampling, adaptive fast forward approaches \cite{Petrovic:2005} try to adjust the speed in different segments of the input video. Sparser frame sampling gives higher speed ups in stationary periods, and denser frame sampling gives lower speed ups in dynamic periods. In general, content aware techniques adjust the frame sampling rate based upon the importance of the content in the video. Typical importance measures include motion in the scene, scene complexity, and saliency. None of the aforementioned methods, however, can handle the challenges of egocentric videos, as we describe next.

Borrowing the terminology of \cite{us_egoseg}, we note that when the camera wearer is ``stationary" (e.g, sitting or standing in place), head motions are less frequent and pose no challenge to traditional fast-forward and stabilization techniques. Therefore, in this paper we focus only on cases when the camera wearer is ``in transit" (e.g, walking, cycling, driving, etc), and often with substantial camera shake.

Kopf \etal \cite{hyperlapse} recently proposed to generate hyperlapse egocentric videos by $3$D reconstruction of the input camera path. A smoother camera path is calculated, and new frames are rendered for this new path using the frames of the original video. Generated video is very impressive, but it may take hours to generate minutes of hyperlapse video. Joshi \etal \cite{rt-hyperlapse} proposed to replace $3$D reconstruction by smart sampling of the input frames. They bias the frame selection in favor of the forward looking frames, and drop frames that might introduce shake.

We model frame sampling as an energy minimization problem. A video is represented as a directed acyclic graph whose nodes correspond to input video frames. The weight of an edge between nodes corresponding to frames $t$ and $t+k$ indicates how ``stable" the output video will be if frame $t+k$ will immediately follow frame $t$. The weights also indicate if the sampled frames give the desired playback speed. Generating a stable fast forwarded video becomes equivalent to finding a shortest path in this graph. We keep all edge weights non-negative, and note that there are numerous polynomial time algorithms for finding a shortest path in such graphs. The proposed frame sampling approach, which we call \egosns, was initially introduced in \cite{egosampling}. We show that sequences produced with \egos are more stable and easier to watch compared to traditional fast forward methods.

\begin{figure}[t]
   \centering
    \includegraphics[width=1\linewidth]{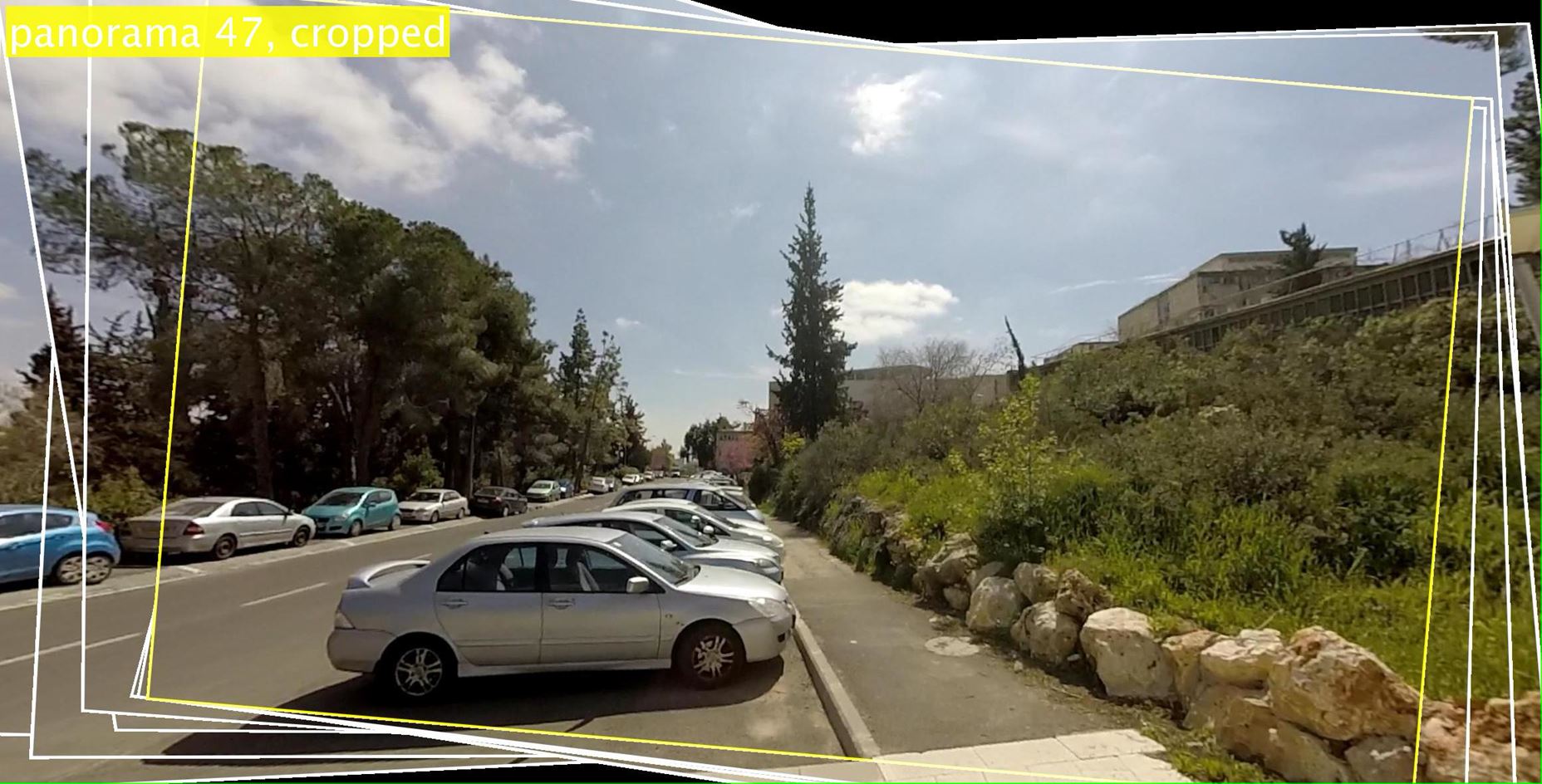}
    \caption{An output frame produced by the proposed \panohns. We collect frames looking into different directions from the video and create mosaics around each frame in the video. These mosaics are then sampled to meet playback speed and video stabilization requirements. Apart from fast forwarded and stabilized, the resulting video now also has wide field of view. The white lines mark the different original frames. The proposed scheme turns the problem of camera shake present in egocentric videos into a feature, as the shake helps increasing the field of view.}
    \label{fig:output_panorama}
\end{figure}

Frame sampling approaches like \egos described above, as well as \cite{hyperlapse, rt-hyperlapse}, drop frames to give a stabilized video, with a potential loss of important information. In addition, a stabilization post-processing is commonly applied to the remaining frames, a process which reduces the field of view. We propose an extension of \egosns, in which instead of dropping unselected frames, these frames are used to increase the field of view of the output video. We call the proposed approach \panohns. Fig.~\ref{fig:output_panorama} shows a frame from an output \panoh generated with our method. \panoh video is easier to comprehend than \cite{rt-hyperlapse} because of its increased field of view. \panoh can also be extended to handle multiple egocentric videos recorded by a groups of people walking together. Given a set of egocentric videos captured at the same scene, \panoh can generate a stabilized panoramic video using frames from the entire set. The combination of multiple videos into a \panoh enables to consume the videos even faster.

The contributions of this work are as follows: i) The generated wide field-of-view, stabilized, fast forward  videos are easier to comprehend than only stabilized or only fast forward videos. ii) The technique is extended to combine together multiple egocentric video taken at the same scene.


The rest of the paper is organized as follows. Relevant related work is described in Sect.~\ref{sec:related_work}. The \egos framework is briefly described in Sect.~\ref{sec:sampling_framework}. In Sect.~\ref{sec:single_input_method} and Sect.~\ref{sec:multiple_input_method} we introduce the generalized \panoh for single and multiple videos, respectively. We report our experiments in Sect.~\ref{sec:experiments}, and conclude in Sect.~\ref{sec:conclusion}.

\begin{figure*}[t]
	\centering
	\includegraphics[width=1\textwidth]{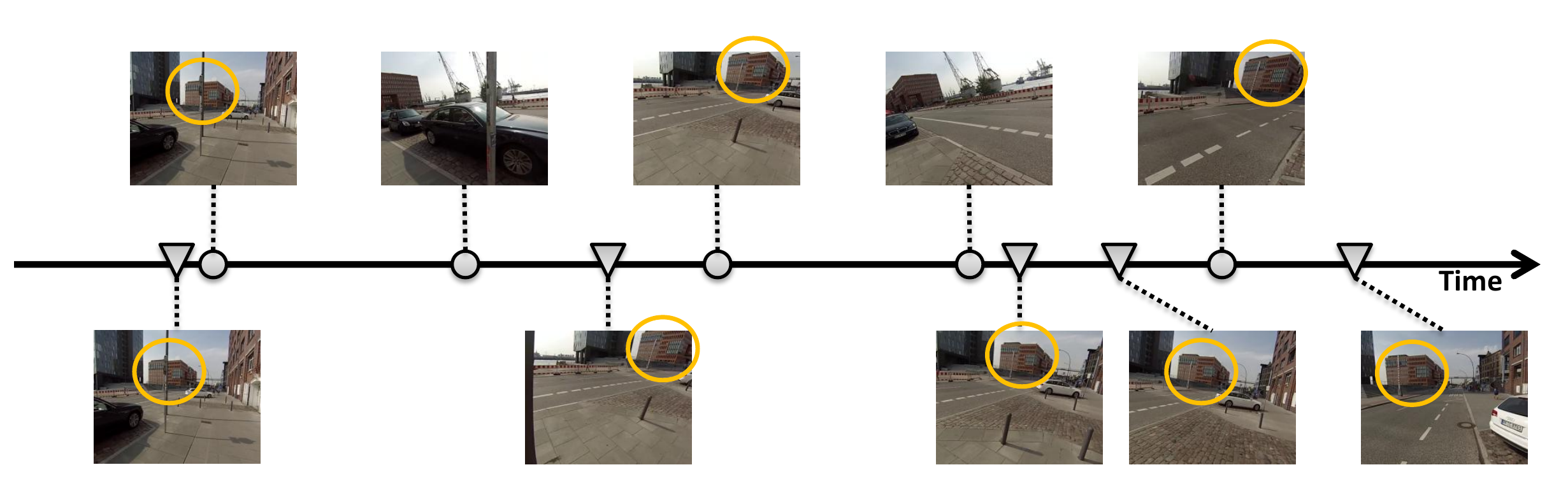}
    \caption{Representative frames from the fast forward results on `Bike2' sequence \cite{hyperlapse-dataset}. The camera wearer rides a bike and prepares to cross the road.
\underline{Top row:} uniform sampling of the input sequence leads to a very shaky output as the camera wearer turns his head sharply to the left and right before crossing the road.
\underline{Bottom row:} \egos prefers forward looking frames and therefore samples the frames non-uniformly so as to remove the sharp head motions. The stabilization can be visually compared by focusing on the change in position of the building (circled yellow) appearing in the scene. The building does not even show up in two frames of the uniform sampling approach, indicating the extreme shake. Note that the fast forward sequence produced by \egos can be post-processed by traditional video stabilization techniques to further improve the stabilization.}
    \label{fig:result_bike2}
\end{figure*}

\section{Related Work}
\label{sec:related_work}

The related work to this paper can be broadly categorized into four categories.

\subsection{Video Summarization}

Video Summarization methods scan the input video for salient events, and create from these events a concise output that captures the essence of the input video. While video summarization of third person videos has been an active research area, only a handful of these works address the specific challenges of summarizing egocentric videos. In \cite{grauman-important-people,grauman-snap-points}, important keyframes are sampled from the input video to create a story-board summarization. In \cite{grauman-story}, subshots that are related to the same ``story" are sampled to produce a ``story-driven" summary. Such video summarization can be seen as an extreme adaptive fast forward, where some parts are completely removed while other parts are played at original speed. These techniques require a strategy for determining the importance or relevance of each video segment, as segments removed from summary are not available for browsing. 

\subsection{Video Stabilization}

There are two main approaches for video stabilization. While $3$D methods reconstruct a smooth camera path \cite{content_preserving_warps,vid3d_stab_depth}, $2$D methods, as the name suggests, use $2$D motion models followed by non-rigid warps \cite{youtube_stabilizer,subspace_vid_stab,BundledPaths2013,steadyflow,raanan}. As noted by \cite{hyperlapse}, stabilizing egocentric video after regular fast forward by uniform frame sampling, fails. Such stabilization can not handle outlier frames often found in egocentric videos, e.g. frames when the camera wearer looks at his shoe for a second, resulting in significant residual shake present in the output videos.


The proposed \egos approach differs from both traditional fast forward as well as video stabilization. Rather than stabilizing outlier frames, we prefer to skip them. However, traditional video stabilization algorithms \cite{youtube_stabilizer, subspace_vid_stab, BundledPaths2013, steadyflow, raanan} can be applied as post-processing to our method, to further stabilize the results.

Traditional video stabilization crop the input frames to create stable looking output with no empty region at the boundaries. In attempt to reduce the cropping, Matsushita \etal \cite{eyalofek_stab} suggest to perform inpainting of the video boundary based on information from other frames.

\subsection{Hyperlapse}

Kopf et al. \cite{hyperlapse} have suggested a pioneering hyperlapse technique to generate stabilized egocentric videos using a combination of $3$D scene reconstruction and image based rendering techniques. A new and smooth camera path is computed for the output video, while remaining close to the input trajectory. The results produced are impressive, but may be less practical because of the large computational requirements. In addition, $3$D recovery from egocentric video may often fail. A similar paper to our \egos approach, \cite{rt-hyperlapse} avoids $3$D reconstruction by posing hyperlapse as frame sampling, and can even be performed in real time.

Sampling-based hyperlapse such as \egos proposed by us or \cite{rt-hyperlapse}, bias the frame selection towards forward looking views. This selection has two effects: (i) The information available in the skipped frames, likely looking sideways, is lost; (ii) The cropping which is part of the subsequent stabilization step, reduces the field of view. We propose to extend the frame sampling strategy by \panohns, which uses the information in the side looking frames that are discarded by frame sampling.

\subsection{Multiple Input Videos}

The state of art hyperlapse techniques address only a single egocentric video. For curating multiple non-egocentric video streams, Jiang and Gu \cite{jiang2015video_stitching} suggested spatial-temporal content-preserving warping for stitching multiple synchronized video streams into a single panoramic video. Hoshen \etal \cite{yedid_curation} and Arev \etal  \cite{arev_auto_social_cams} produce a single output stream from multiple egocentric videos viewing the same scene.  This is done by selecting only a single input video, best representing each time period. In both the techniques, the criterion for selecting the video to display requires strong assumptions of what is interesting and what is not.

We propose \panoh in this paper, which supports multiple input videos, by fusing input frames from multiple videos into a single output frame having a wide field of view.

\section{\egosns} \label{sec:sampling_framework}

The key idea in this paper is to generate a stable fast forwarded output video by selecting frames from the input video having similar forward viewing direction, which is also the direction of the wearer's motion. Fig.~\ref{fig:result_bike2} intuitively describes this approach. This approach works well for forward moving cameras.  Other motion directions, e.g. cameras moving sideways, can be accelerated only slightly before becoming hard to watch.

As a measure for forward looking direction, we find the Epipolar point between all pairs of frames, $I_t$ and $I_{t+k}$, where $k \in [1,\tau]$, and $\tau$ is the maximum allowed frame skip. Under the assumption that the camera is always translating (recall that we focus only on wearer's in ``transit' state), the displacement direction between $I_t$ and $I_{t+k}$ can be estimated from the fundamental matrix $F_{t,t+k}$ \cite{hartley_book}. We prefer using frames whose epipole is closest to the center of the image.

Recent V-SLAM approaches such as \cite{lsdslam,svo} provide camera ego-motion estimation and localization in real-time. However, we found that the fundamental matrix computation can fail frequently when $k$ (temporal separation between the frame pair) grows larger. As a fallback measure, whenever the fundamental matrix computation breaks, we estimate the direction of motion from the FOE of the optical flow. We do not compute the FOE from the instantaneous flow, but from integrated optical flow as suggested in \cite{us_egoseg} and computed as follows: (i) We first compute the sparse optical flow between all consecutive frames from frame $i$ to frame $j$. Let the optical flow between frames $t$ and $t+1$ be denoted by $g_t(x,y)$ and
$G_{i,j}(x,y) =  \frac{1}{k} \sum_{t=i}^{j-1} g_t(x,y)$. The FOE is computed from $G_{i,j}$ as suggested in \cite{technion-foe}, and is used as an estimate of the direction of motion.





\begin{figure}[t]
    \centering
    \includegraphics[width=1\linewidth]{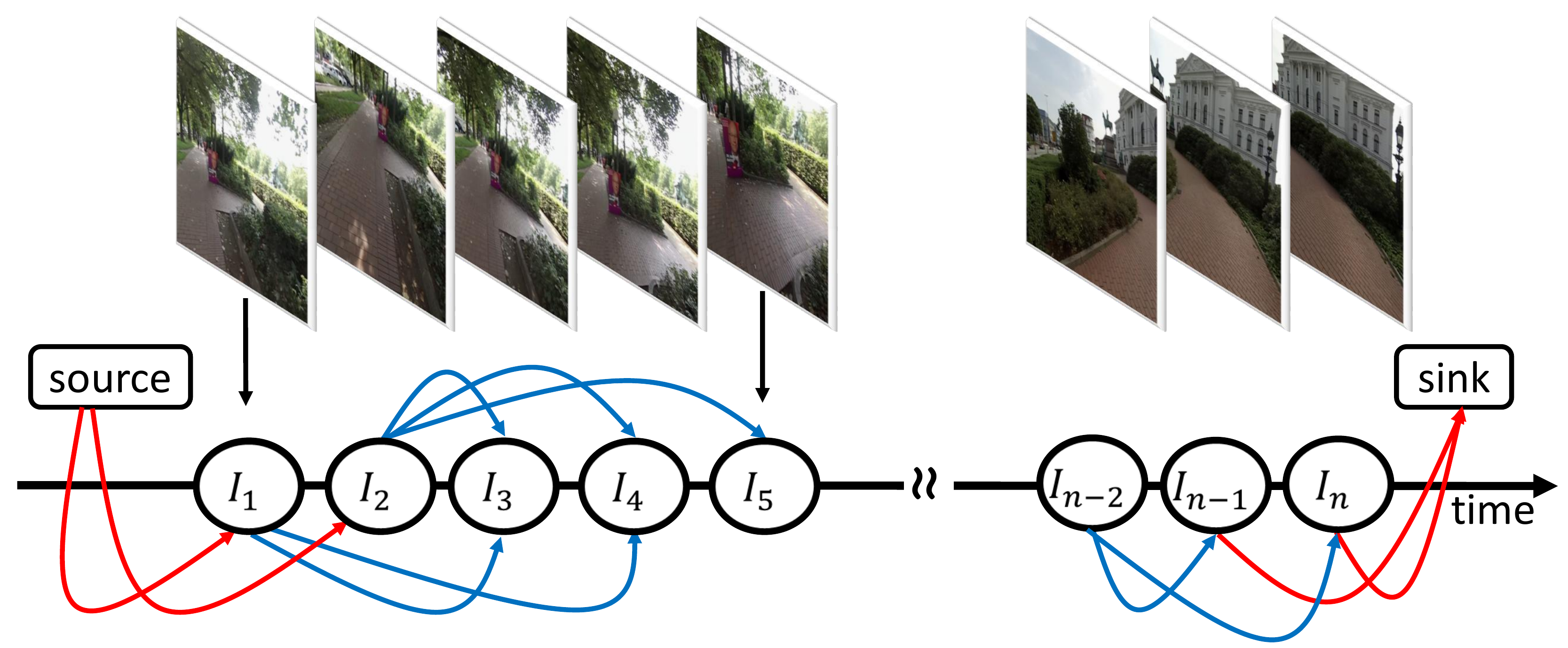}
    \caption{We formulate the joint fast forward and video stabilization problem as finding a shortest path in a graph constructed as shown. There is a node corresponding to each frame. The edges between a pair of frames $(i,j)$ indicate the penalty for including a frame $j$ immediately after frame $i$ in the output (please refer to the text for details on the edge weights). The edges between the source/sink and the graph nodes allow to skip frames from start and end. The frames corresponding to nodes along the shortest path from the source to the sink are included in the output video.}
    \label{fig:first_order_graph}
\end{figure}

\begin{figure*}[t]
    \centering
	\includegraphics[width=0.8\linewidth]{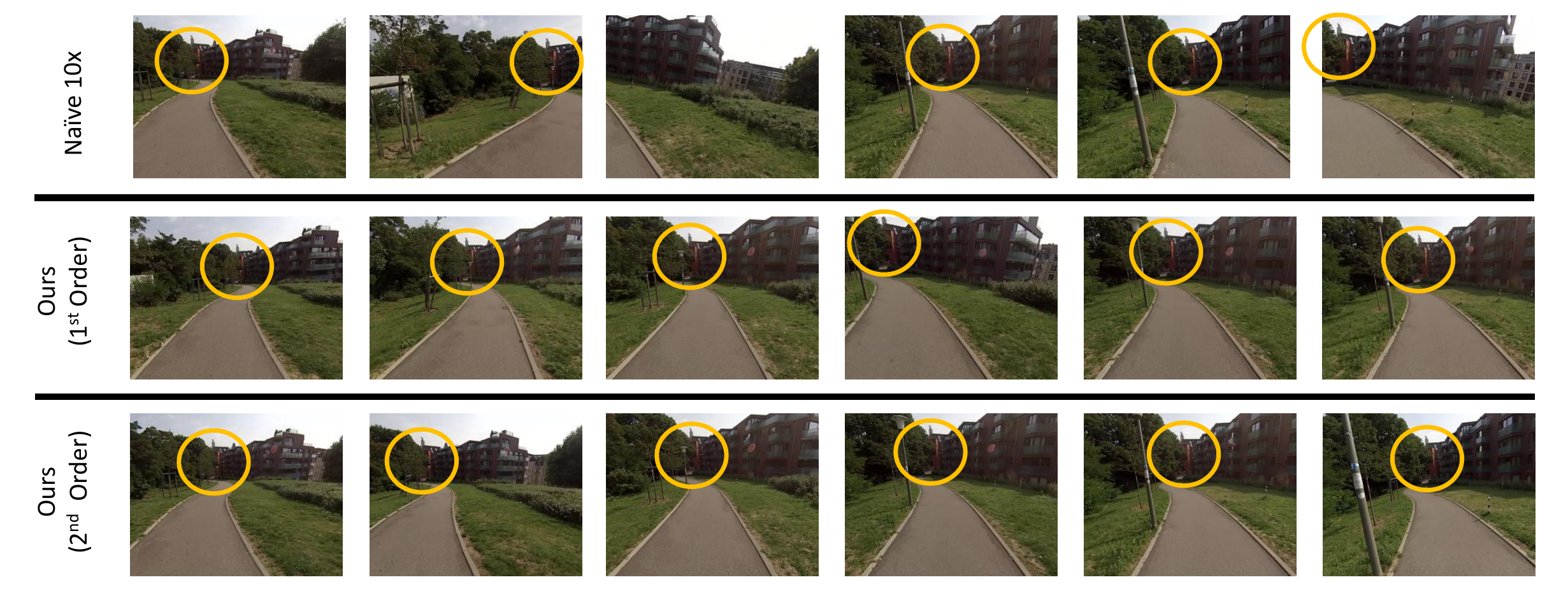}
    \caption{Comparative results for fast forward from na\"{\i}ve uniform sampling (first row), \egos using first order formulation (second row) and using second order formulation (third row). Note the stability in the sampled frames as seen from the tower visible far away (circled yellow). The first order formulation leads to a more stable fast forward output compared to na\"{\i}ve uniform sampling. The second order formulation produces even better results in terms of visual stability.}
    \label{fig:res_ff_comparison}
\end{figure*}


\subsection{Graph Representation}
\label{sec:graph}

We model the joint fast forward and stabilization of egocentric video as graph energy minimization. The input video is represented as a graph, with a node corresponding to each frame in the video. There are weighted edges between every pair of graph nodes, $i$ and $j$, with weight proportional to our preference for including frame $j$ right after $i$ in the output video. There are three components in this weight:

\begin{enumerate}
    \item Shakiness Cost ($S_{i,j}$): This term prefers forward looking frames. The cost is proportional to the distance of the computed motion direction (Epipole or FOE) designated by $(x_{i,j}, y_{i,j})$ from the center of the image $(0,0)$:
\begin{equation}
	{S}_{i,j}=\|(x_{i,j}, y_{i,j})\|
\end{equation}
    \item Velocity Cost ($V_{i,j}$): This term controls the playback speed of the output video. The desired speed is given by the desired magnitude of the optical flow, $K_{flow}$, between two consecutive output frames.
\begin{equation}
    V_{i,j}=(\sum_{x,y} G_{i,j}(x,y) - K_{flow})^2
\end{equation}

    \item Appearance Cost ($C_{i,j}$): This is the Earth Mover's Distance (EMD) \cite{emd} between the color histograms of frames $i$ and $j$. The role of this term is to prevent large visual changes between frames. A quick rotation of the head or dominant moving objects in the scene can confuse the FOE or epipole computation. This term acts as an anchor in such cases, preventing the algorithm from skipping a large number of frames.
\end{enumerate}

The overall weight of the edge between nodes (frames) $i$ and $j$ is given by:
\begin{equation}
\mathcal{W}_{i,j}=\alpha\cdot\mathcal{S}_{i,j}+\beta\cdot V_{i,j}+\gamma\cdot C_{i,j},
\end{equation}
where $\alpha$, $\beta$ and $\gamma$ represent the relative importance of various costs in the overall edge weight.

With the problem formulated as above, sampling frames for stable fast forward is done by finding a shortest path in the graph. We add two auxiliary nodes, a \emph{source} and a \emph{sink} in the graph to allow skipping some frames from start or end. To allow such skip, we add zero weight edges from start node to the first $D_{start}$ frames and from the last $D_{end}$ nodes to sink. We then use Dijkstra's algorithm \cite{dijkstra} to compute the shortest path between source and sink. The algorithm does the optimal inference in time polynomial in the number of nodes (frames). Fig.~\ref{fig:first_order_graph} shows a schematic illustration of the proposed formulation.


\subsection{Second Order Smoothness}

\begin{figure}[t]
    \centering
    \includegraphics[width=1\linewidth]{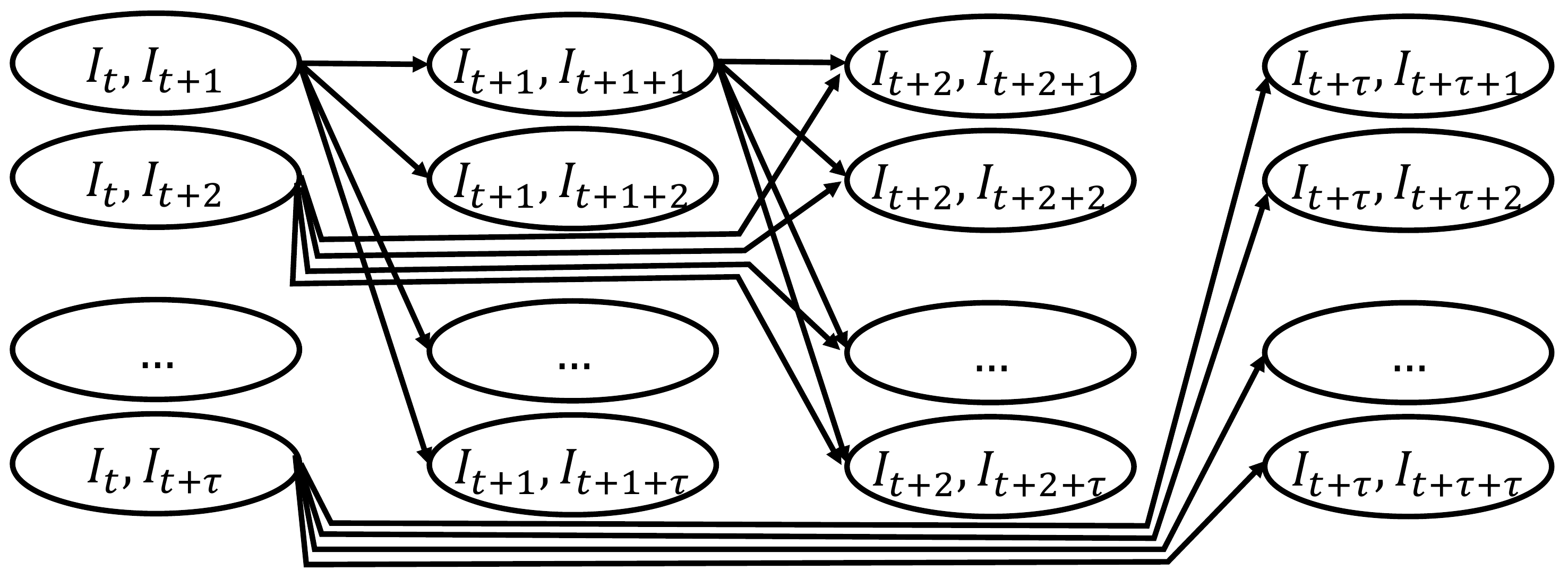}
    \caption{The graph formulation, as described in Fig.~\ref{fig:first_order_graph}, produces an output which has almost forward looking direction. However, there may still be large changes in the epipole locations between two consecutive frame transitions, causing jitter in the output video.
To overcome this we add a second order smoothness term based on triplets of output frames.
    Now the nodes correspond to pairs of frames, instead of single frames in the first order formulation described earlier. There are edges between frame pairs $(i,j)$ and $(k,l)$,  if $j=k$. The edge reflects the penalty for including frame triplet $(i,k,l)$ in the output. Edges from source and sink to graph nodes (not shown in the figure) are added in the same way as in the first order formulation to allow skipping frames from start and end.}
    \label{fig:second_order_graph}
\end{figure}

The formulation described in the previous section prefers to select forward looking frames, where the epipole is closest to the center of the image. With the proposed formulation, it may so happen that the epipoles of the selected frames are close to the image center but on the opposite sides, leading to a jitter in the output video. In this section we introduce an additional cost element: stability of the location of the epipole. We prefer to sample frames with minimal variation of the epipole location.

To compute this cost, nodes now represent two frames, as can be seen in Fig.~\ref{fig:second_order_graph}. The weights on the edges depend on the change in epipole location between one image pair to the successive image pair. Consider three frames $I_{t_1}$, $I_{t_2}$ and $I_{t_3}$. Assume the epipole between $I_{t_i}$ and $I_{t_j}$ is at pixel $(x_{ij}, y_{ij})$. The second order cost of the triplet (graph edge)  $(I_{t_1},I_{t_2},I_{t_3})$, is proportional to $\|(x_{23}-x_{12}, y_{23}-y_{12})\|$.

This second order cost is added to the previously computed shakiness cost.
The graph with the second order smoothness term has all edge weights non-negative and the running-time to find an optimal solution to shortest path is linear in the number of nodes and edges, i.e. $O(n\tau^2)$. In practice, with $\tau=100$, the optimal path was found in all examples in less than 30 seconds. Fig.~\ref{fig:res_ff_comparison} shows results obtained from both first order and second order formulations.


\section{\panoh of a Single Video}
\label{sec:single_input_method}

\begin{figure}[t]
   \centering
    \includegraphics[width=1\linewidth]{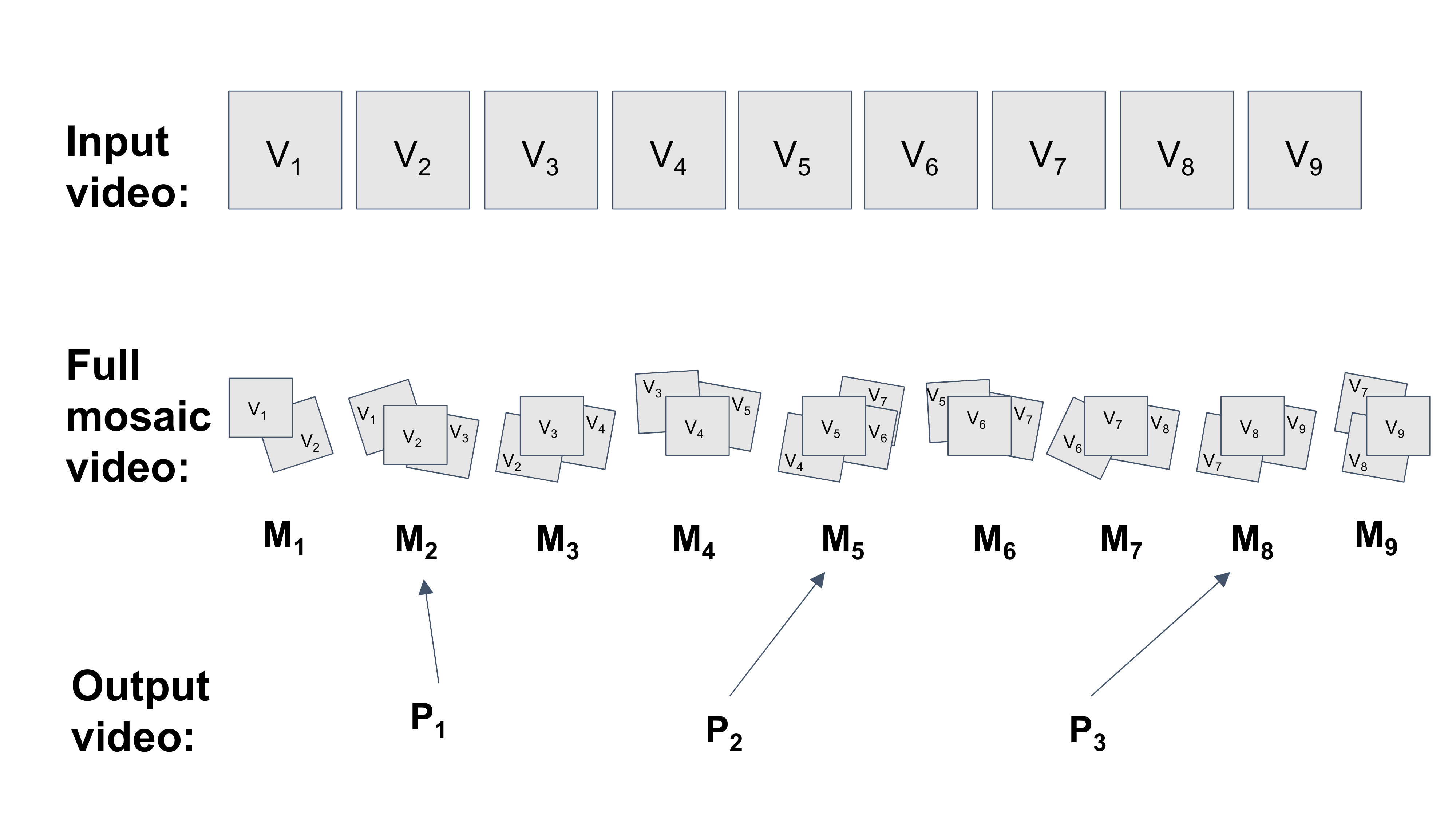}
    \caption{\panoh creation. At the first step, for each input frame $v_i$ a mosaic $M_i$ is created from frames before and after it. At the second stage, a \panoh video $P_i$ is sampled from $M_i$ using sampled hyperlapse methods such as \cite{rt-hyperlapse} or \egosns.}
    \label{fig:mosaic-sample-after-construct}
\end{figure}

Sampling based hyperlapse techniques (hereinafter referred to as `sampled hyperlapse'), such as \egosns, or as given in \cite{rt-hyperlapse}, drop many frames for output speed and stability requirements. Instead of simply skipping the unselected frames which may contain important events, we suggest ``\panohns", which uses all the frames in the video for building a panorama around selected frames.

\subsection{Creating Panoramas}
\label{sec:finding-local-centers}

For efficiency reasons, we create panoramas only around carefully selected central frames. The panorama generation process starts with the chosen frame as the reference frame.  This is a common approach in mosaicing that reference view for the panorama should be ``the one that is geometrically most central" (\cite{szeliski2006image}, p. 73). In order to choose the best central frame, we take a window of $\omega$ frames around each input frame and track feature points through this temporal window.

Let $f_{i,t}$ be the displacement of feature point $i \in \{1 \ldots n\}$ in frame $t$ relative to its location in the first frame of the temporal window. The displacement of frame $t$ relative to the first frame is defined as: 
\begin{equation}
pos_{t}=\frac{1}{n}\sum\limits_{i=1}^n f_{i,t}
\end{equation}
and the central frame is
\begin{equation}
\bar{t}=\arg\!\min_{t}\{\|pos_{t}-\frac{1}{\omega}\sum\limits_{s=1}^\omega pos_{s}\|\}
\end{equation}

Given the natural head motion alternately to the left and right, the proposed frame selection strategy prefers forward looking frames as central frames.

After choosing the central frame, we align all the frames in the $\omega$ window with the central frame using a homography, and stitch the panorama using the ``Joiners" method \cite{zelnik2007automating}, such that central frames are on top and peripheral frames are at the bottom. More sophisticated stitching and blending, e.g. min-cut and Poisson blending, can be used to improve the appearance of the panorama, or dealing with moving objects, etc.

\begin{figure}[t]
   \centering
    \includegraphics[width=0.8\linewidth]{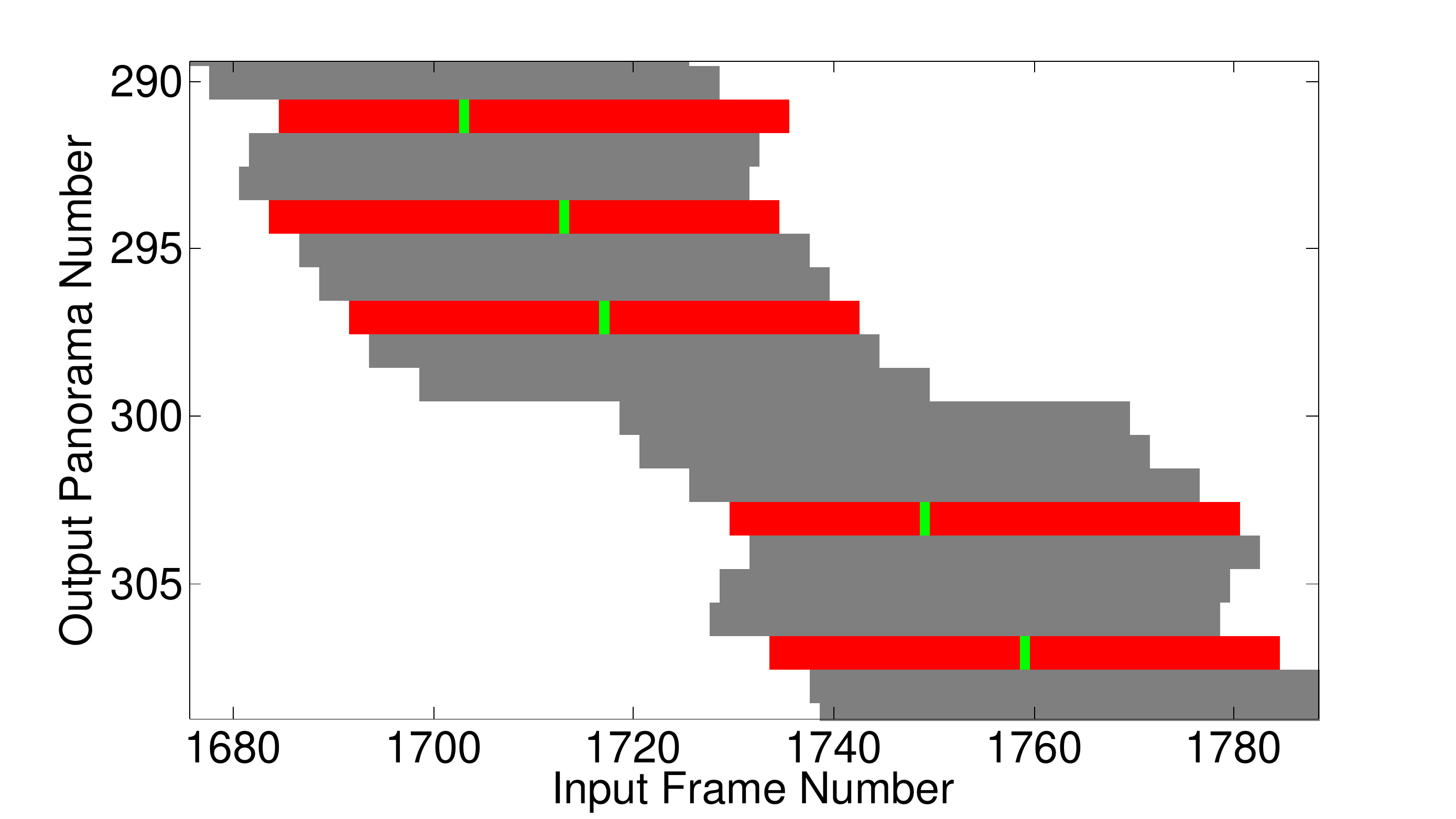}
    \caption{An example for mapping input frames to output panoramas from sequence `Running'. Rows represent generated panoramas, columns represent input frames. Red panoramas were selected for \panohns, and gray panoramas were not used. Central frames are indicated in green.  }
    \label{fig:input_frame_to_panorama}
\end{figure}

\subsection{Sampling Panoramas} \label{sec:sampling-panoramas}

After generating panoramas corresponding to different central frames, we sample a subset of panoramas for the hyperlapse video. The sampling strategy is similar to the process described in Section \ref{sec:sampling_framework}, with the nodes now corresponding to panoramas and the edge weight representing the cost of the  transition from panorama $p$ to panorama $q$, defined as follows:
\begin{equation}
W_{p,q} = \alpha \cdot S_{p,q} + \beta \cdot V_{p,q} + \gamma \cdot FOV_{p}.
\label{eq:cost}
\end{equation}
Here, the shakiness $S_{p,q}$ and the velocity $V_{p,q}$ are measured between the central frames of the two panoramas. $FOV_p$ denotes the size of the panorama $p$, and is counted as the number of pixels painted by all frames participating in that panorama. We measure it by warping the four corners of each frame to determine the area that will be covered by the actual warped images. In the end, we run the shortest path algorithm to select the sampled panoramas as described in the previous section.

Fig.~\ref{fig:input_frame_to_panorama} shows the participation of input frames in the panoramas for one of the sample sequence. We show in gray the candidate panoramas before sampling, and the finally selected panoramas are shown in red. The span of each row shows the frames participating in each panorama.

\subsection{Stabilization}

In our experiments we performed minimal alignment between panoramas, using only a rigid transformation between the central frames of the panoramas. When feature tracking was lost we placed the next panorama at the center of the canvas and started tracking from that frame. Any stabilization algorithm may be used as a post processing step for further fine detail stabilization. Since video stabilization reduces the field of view to be only the common area seen in all frames, starting with panoramic images mitigates this effect.

\subsection{Cropping}

Panoramas are usually created on a canvas much larger than the size of the original video, and large parts of the canvas are not covered with any of the input images. In our technique, we applied a moving crop window on the aligned panoramas. The crop window was reset whenever the stabilization was reset. In order to get smooth window movement, while containing as many pixels as possible we find crop centers $cr_{i}$ which minimize the following energy function:
\begin{equation}
E = \sum \|cr_{i}-m_{i} \|^{2} + \lambda \sum \|cr_{i}-\frac{cr_{i-1} + cr_{i+1}}{2} \|^{2},
\end{equation}
where $m_{i}$ is the center of mass of the $i^\text{th}$ panorama. This can be minimized by solving the sparse set of linear equations given by the derivatives:
\begin{equation}
cr_{i} = \frac{\lambda ( cr_{i-1} + cr_{i+1} ) + m_{i} }{2 \lambda + 1}
\end{equation}
The crop size is dependent on the camera movement and on $\lambda$. Larger $\lambda$ will favor less movement of the crop window, and in order to keep it in the covered part of the canvas it will get smaller.

\subsection{Removing Lens Distortion}

\begin{figure}[t]
   \centering
    \includegraphics[width=1\linewidth]{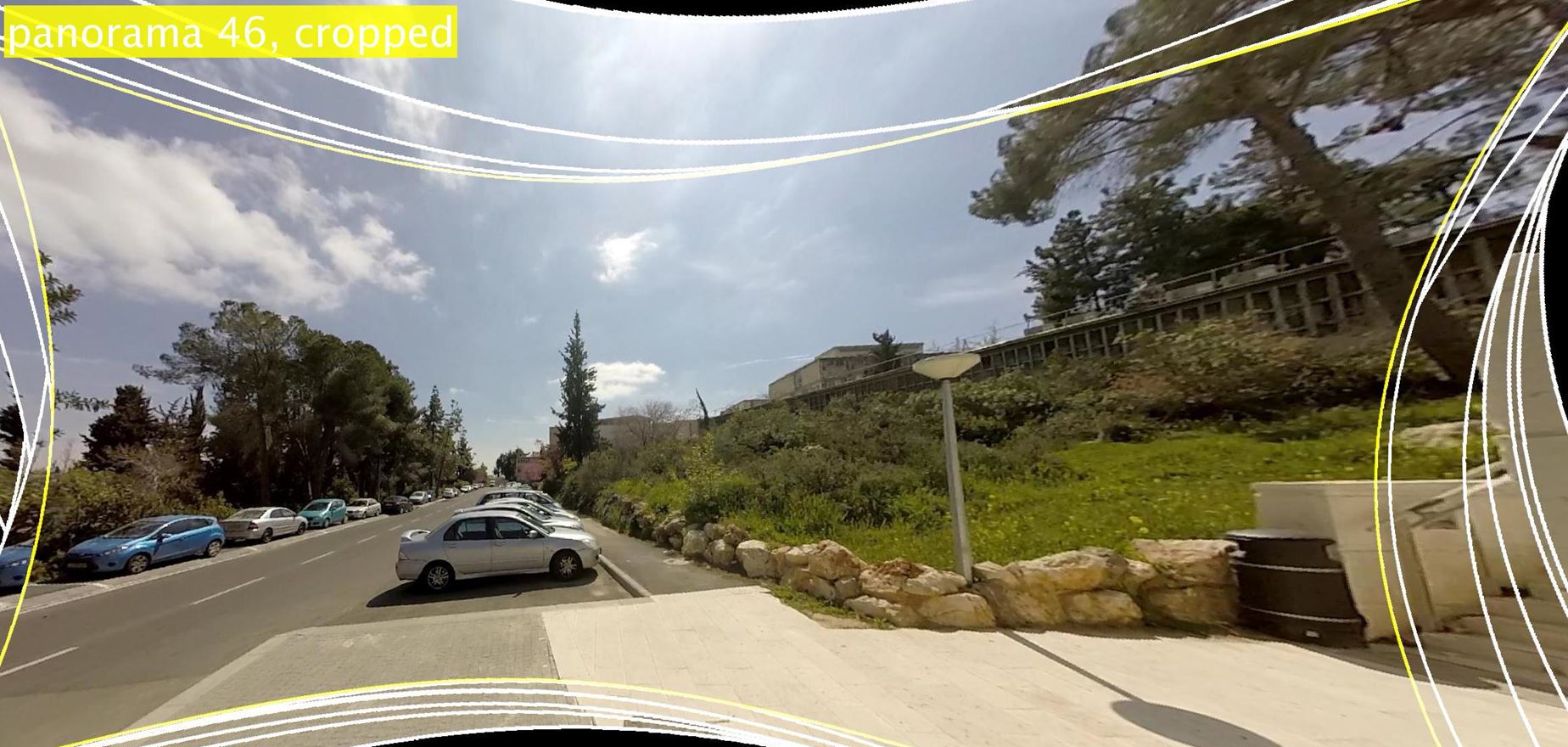}
    \caption{The same scene as in Fig.~\ref{fig:output_panorama}. The frames were warped to remove lens distortion, but were not cropped. The mosaicing was done on the uncropped frames. Notice the increased FOV compared to the panorama in \ref{fig:output_panorama}.}
    \label{fig:mosaic-lens-distortion}
\end{figure}

\begin{algorithm}[t]
\label{alg_single}
 \KwData{Single video}
 \KwResult{\panohns}
 \For{every temporal window}{
	find the central frame of the window\;
  }
  \For{every panorama candidate with center $c$}{
  	\For{each frame $f$ participating in the panorama}{
		Calculate the transformation between $f$ and $c$\;
        Calculate the cost for shakiness, FOV and velocity\;
    }
  }
  Choose panoramas for the output using shortest path in graph algorithm\;
  Construct the panoramas\;
  Stabilize and crop\;
 \caption{Single video \panohns}
\end{algorithm}

We use the method of \cite{ocamlib} to remove lens distortion. Usually, frames are cropped after the lens distortion removal to a rectangle containing only valid pixels. However, in the case of panoramas, the cropping may be done after stitching the frames. This results in even larger field of view. An example of a cropped panoramic image after removal of lens distortion is given in Figure \ref{fig:mosaic-lens-distortion}.

We list the steps to generate \panoh in Algorithm \ref{alg_single}.

\section{\panoh of Multiple Videos}
\label{sec:multiple_input_method}

\panoh can be extended naturally to multiple input videos, as we show in this section.

\subsection{Correspondence Across Videos}

For multi-video hyperlapse, we first find corresponding frames in all other videos, for every frame in each video. We define as corresponding frame, the frame having the largest region of overlap, measured by the number of matching feature points between the frames. Any pair of frames with less than 10 corresponding points is declared as non-overlapping. We used coarse-to-fine strategy, starting from approximate candidates with skip of 10 frames between each pair of matched images to find a searching interval, and then zeroing on the largest overlapping frame in that interval. It may be noted that, some frames in one video may not have corresponding frame in the second video. Also note that the corresponding frame relationship is not symmetric.

We maintain temporal consistency in the matching process. For example, assuming $x'$ and $y'$ are the corresponding frame numbers in the second video for frame numbers $x$ and $y$ in the first video. If $x<y$, then we drop the match $y,y'$ if $x'>y'$.

\begin{figure}[t]
   \centering
    \includegraphics[width=1\linewidth]{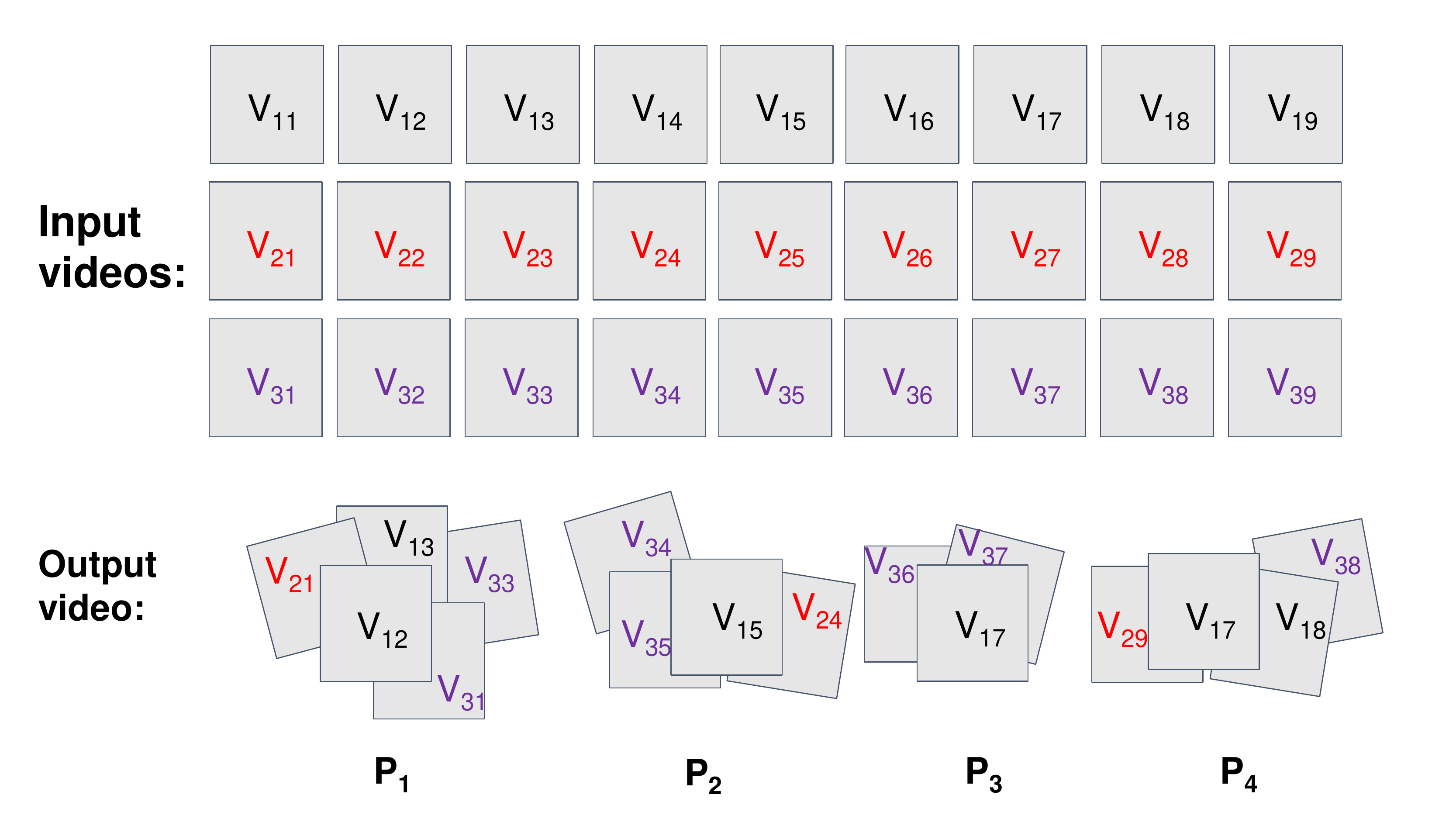}
    \caption{Creating a multi-video \panohns. The first three rows indicate three input videos with frames labeled $V_{ij}$. Each frame $P_{i}$ in the output panoramic video is constructed by mosaicing one or more of the input frames, which can originate from any input video.}
    \label{fig:mosaic-multi-video}
\end{figure}

\begin{figure}[t]
   \centering
    \includegraphics[width=1\linewidth]{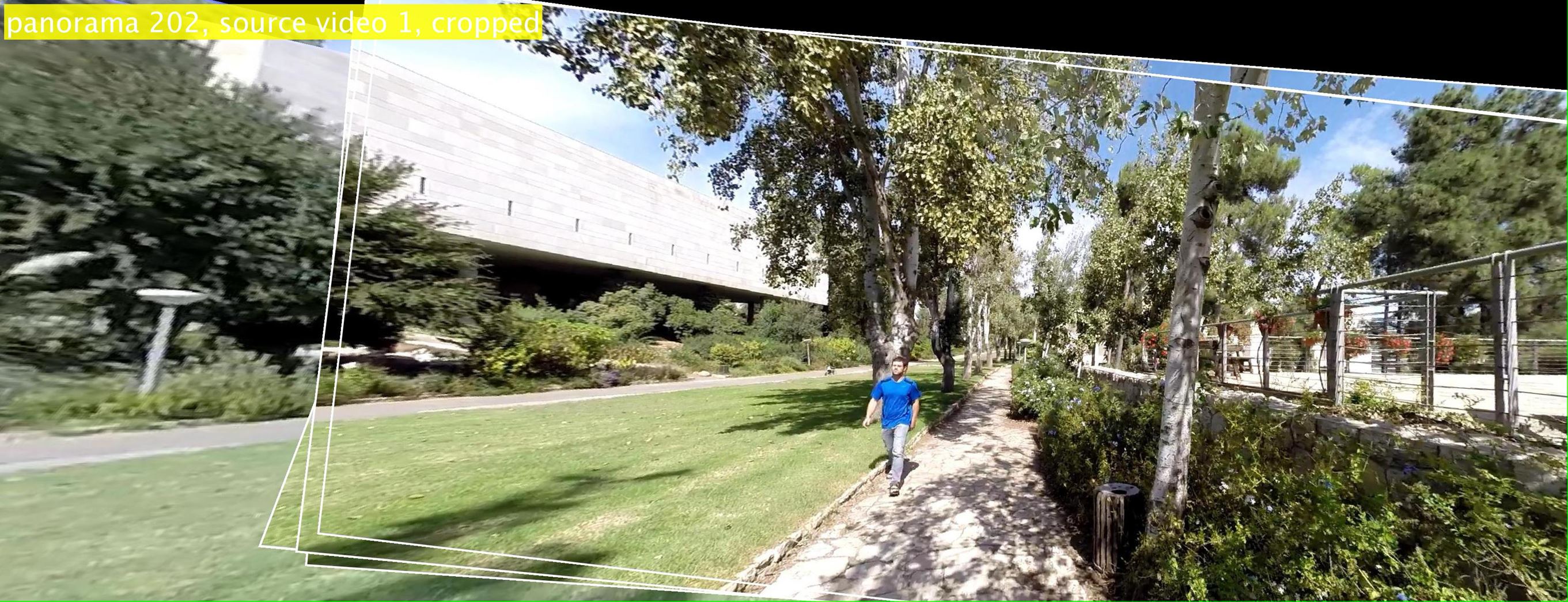}
    \caption{A multi-video output frame. All rectangles with white borders are frames from the same video, while the left part is taken from another. Notice the enlarged field of view resulting from using frames from multiple videos.}
    \label{fig:multi-video}
\end{figure}

\begin{algorithm}[t]
\label{alg_multi}
 \KwData{Multiple videos}
 \KwResult{\panohns}
\textbf{Preprocess:} temporally align videos (if necessary)\;
    			calculate homographies between matching frames in different videos\;
 \For{each video}{
 	Find central frames and calculate cost similar to the single video case;
 }
  Calculate cross-video cost \;
  Choose panoramas for the output using shortest path in graph algorithm\;
  \For{each panorama with center $c$}{
  	  \For {every frame $f$ from $c$'s video participating in the panorama}{
	  warp $f$ towards $c$\;
		\For {frames $f'$ aligned with $f$ in other videos}{
			warp $f'$ towards $c$ using chained homography $f'$-$f$-$c$\;
         }
      }
	  Construct the panoramas;
  }
  Stabilize and crop\;
\caption{Multi video \panohns}
\end{algorithm}

\subsection{Creation of Multi-Video Panorama}

Once the corresponding frames have been identified, we initiate the process of selecting central frames. This process is done independently for each video as described in Sec.~\ref{sec:single_input_method} with the difference that for each frame in the temporal window $\omega$, we now collect all corresponding frames from all the input videos. For example, in an experiment with $n$ input videos, up to $(n \cdot |\omega|)$ frames may participate in each central frame selection and mosaic generation process. The process of panorama creation is repeated for all temporal windows in all input videos. Fig.~\ref{fig:mosaic-multi-video} outlines the relation between the \panoh and the input videos. Note that the process of choosing central frames for each camera ensures that the stabilization achieved in multi-video \panoh is similar to the one that would have been achieved if there were only a single camera. The mosaic creation can only increase the sense of stabilization because of increased field of view.

\subsection{Sampling}

After creating panoramas in each video, we perform a sampling process similar to the one described in Sec.~\ref{sec:sampling-panoramas}. The difference being that the candidate panoramas for sampling come from all the input videos. The graph creation process is the same with the nodes now corresponding to panoramas in all the videos. For the edge weights, apart from the costs as mentioned in the last section, we insert an additional term called \emph{cross-video} penalty. Cross-video terms add a switching penalty, if in the output video there is a transition from panorama with central frame from one video to a panorama with central frame that comes from some other video. Note that the FOE stabilization cost in the edge weight aims to align the viewing angles of two (or three) consecutive frames in the output video and is calculated similarly irrespective of whether the input frames originated from single or multiple videos.

The shortest path algorithm then runs on the graph created this way and chooses the panoramic frames from all input videos. We show a sample frame from one of the output videos generated by our method in Fig.~\ref{fig:multi-video}. Algorithm \ref{alg_multi} gives the pseudocode for our algorithm.

It may be noted that the proposed scheme samples the central frames judiciously on the basis of \egos, with the quality of the chosen output mosaics being a part of the optimization. This is not equivalent to generating mosaics from individual frames and then generating the stabilized output, in the same way as in the case of single video scenario, fast forward followed by stabilization is not equivalent to \egosns.

\section{Experiments} \label{sec:experiments}

In this section we give implementation details and show the results for \egos as well as \panohns. We have used publicly available sequences \cite{hyperlapse-dataset, youtube-running1, youtube-driving2, ego_social} as well as our own videos for the demonstration. The details of the sequences are given in Table \ref{tb:ff_sequences}. We used a modified (faster) implementation of \cite{us_egoseg} for the LK \cite{lk} optical flow estimation. We use the code and calibration details given by \cite{hyperlapse} to correct for lens distortion in their sequences. Feature point extraction and fundamental matrix recovery is performed using VisualSFM \cite{visualsfm}, with GPU support. The rest of the implementation (FOE estimation, energy terms and shortest path etc.) is in Matlab. All the experiments have been conducted on a standard desktop PC.

\renewcommand{\tabcolsep}{0.3cm}
\begin{table}[t]
    \caption{Sequences used for the fast forward algorithm evaluation. All sequences were shot in 30fps, except 'Running' which is 24fps and 'Walking11' which is 15fps.}
    \label{tb:ff_sequences}
    \centering
    \begin{tabular}{lccccc}
        \toprule[1.5pt]
        \specialcell{\bf \tabletext Name}  &
        \specialcell{\bf \tabletext Src}  &
        \specialcell{\bf \tabletext Resolution} &
        \specialcell{\bf \tabletext Num\\ \bf \tabletext Frames} \\ \midrule

	\tabletext Walking1		& \specialcell{\tabletext \cite{hyperlapse-dataset}} &\specialcell{\tabletext $1280$x$960$}		& \specialcell{\tabletext $17249$} \\

	\tabletext Walking2		& \specialcell{\tabletext \cite{yedid_biometrics}} &\specialcell{\tabletext $1920$x$1080$}	& \specialcell{\tabletext $2610$} \\

	\tabletext Walking3		& \specialcell{\tabletext \cite{yedid_biometrics}} &\specialcell{\tabletext $1920$x$1080$}	& \specialcell{\tabletext $4292$}  \\

	\tabletext Walking4		& \specialcell{\tabletext \cite{yedid_biometrics}} &\specialcell{\tabletext $1920$x$1080$}	& \specialcell{\tabletext $4205$} \\

	\tabletext Walking5		& \specialcell{\tabletext \cite{ego_social}} &\specialcell{\tabletext $1280$x$720$}	& \specialcell{\tabletext $1000$} \\

	\tabletext Walking6		& \specialcell{\tabletext \cite{ego_social}} &\specialcell{\tabletext $1280$x$720$}	& \specialcell{\tabletext $1000$} \\

	\tabletext Walking7		& \specialcell{\tabletext --} &\specialcell{\tabletext $1280$x$960$}	& \specialcell{\tabletext $1500$} \\

	\tabletext Walking8		& \specialcell{\tabletext --} &\specialcell{\tabletext $1920$x$1080$}	& \specialcell{\tabletext $1500$} \\

	\tabletext Walking9		& \specialcell{\tabletext \cite{yedid_biometrics}} &\specialcell{\tabletext $1920$x$1080$}	& \specialcell{\tabletext $2000$} \\
    	
        \tabletext Walking11	&	\cite{ego_social}	&\specialcell{\tabletext $1280$x$720$}		&	\tabletext $6900$		 \\

        \tabletext Walking12	&	\cite{us_egoseg}	&  	\specialcell{\tabletext $1920$x$1080$} &	\tabletext $8001$	\\

        \tabletext Driving		&	\cite{youtube-driving2}	& 	\specialcell{\tabletext $1280$x$720$ }  		&	\tabletext $10200$  \\ %

        \tabletext Bike1		&	\cite{hyperlapse-dataset}	& \specialcell{\tabletext $1280$x$960$}	&	\tabletext $10786$  \\

        \tabletext Bike2		&	\cite{hyperlapse-dataset}	& \specialcell{\tabletext  $1280$x$960$}	 &	\tabletext $7049$    \\

        \tabletext Bike3		&	\cite{hyperlapse-dataset}	& \specialcell{\tabletext  $1280$x$960$}	 &	\tabletext $23700$  \\

        \tabletext Running		&	\cite{youtube-running1}	& 	\specialcell{\tabletext $1280$x$720$} &	\tabletext $12900$  \\

        \bottomrule[1.5pt] \\
    \end{tabular}
\end{table}

\renewcommand{\tabcolsep}{0.1cm}
\begin{table}
    \caption{Fast forward results with desired speedup of factor $10$ using second-order smoothness. We evaluate the improvement as degree of epipole smoothness in the output video (column $5$). The proposed method gives huge improvement over na\"{\i}ve fast forward in all but one test sequence (see Fig.~\ref{fig:ff-failure-driving} for the failure case). Note that the actual skip (column $4$) can differ a lot from the target in the proposed algorithm.}
    \label{tb:ff}
    \centering
    \begin{tabular}{lccccccc}
        \toprule[1.5pt]
        \specialcell{\bf \tabletext Name}  &\specialcell{\bf \tabletext Input\\ \bf \tabletext Frames} &\specialcell{\bf \tabletext Output \\ \bf \tabletext Frames}  & \specialcell{\bf \tabletext Median \\ \bf \tabletext Skip} & \specialcell{\bf \tabletext Improvement over \\ \bf \tabletext Na\"{\i}ve $10\times$} \\  \midrule

        \tabletext Walking1		& \tabletext $17249$  &	  \tabletext $931$ &   \tabletext $17$  &  $283\%$ \\
        \tabletext Walking11	&	\tabletext $6900$	&    \tabletext $284$ &    \tabletext $13$ &	$88\%$ \\ 	
        \tabletext Walking12	& 	\tabletext $8001$	& 	 \tabletext $956$ &    \tabletext $4$ & $56\%$	 \\
        \tabletext Driving		&	\tabletext $10200$  & \tabletext $188$ &   \tabletext $48$ & $-7\%$ \\ %
        \tabletext Bike1		&	\tabletext $10786$   &  \tabletext $378$ &  \tabletext $13$ & $235\%$ \\ 
        \tabletext Bike2 &	\tabletext $7049$      &  \tabletext $343$ &    \tabletext $14$	& $126\%$  \\
        \tabletext Bike3		&    \tabletext $23700$  &  \tabletext $1255$ &  \tabletext $12$ & $66\%$ \\
        \tabletext Running		&	\tabletext $12900$   &  \tabletext $1251$ & \tabletext $8$   & $200\%$ \\

        \bottomrule[1.5pt] \\
    \end{tabular}
\end{table}

\subsection{\egosns}

We show results for \egos on $8$ publicly available sequences. For the $4$ sequences for which we have camera calibration information, we estimated the motion direction based on epipolar geometry. We used the FOE estimation method as a fallback when we could not recover the fundamental matrix. For this set of experiments we fix the following weights: $\alpha=1000$, $\beta=200$ and $\gamma=3$. We further penalize the use of estimated FOE instead of the epipole with a constant factor $c=4$. In case camera calibration is not available, we used the FOE estimation method only and changed $\alpha=3$ and $\beta=10$. For all the experiments, we fixed $\tau=100$ (maximum allowed skip). We set the source and sink skip to $D_{start}=D_{end}=120$ to allow more flexibility. We set the desired speed up factor to $10\times$ by setting $K_{flow}$ to be $10$ times the average optical flow magnitude of the sequence. We show representative frames from the output for one such experiment in Fig.\ref{fig:res_ff_comparison}. Output videos from other experiments are given at the project's website: \url{http://www.vision.huji.ac.il/egosampling/}.

\subsubsection{Running times}

The advantage of \egos is in its simplicity, robustness and efficiency. This makes it practical for long unstructured egocentric videos. We present the coarse running time for the major steps in our algorithm below. The time is estimated on a standard Desktop PC, based on the implementation details given above. Sparse optical flow estimation (as in \cite{us_egoseg}) takes 150 milliseconds per frame. Estimating F-Mat (including feature detection and matching) between frame $I_t$ and $I_{t+k}$ where $k\in[1,100]$ takes 450 milliseconds per input frame $I_t$. Calculating second-order costs takes 125 milliseconds per frame. This amounts to total of 725 milliseconds of processing per input frame. Solving for the shortest path, which is done once per sequence, takes up to 30 seconds for the longest sequence in our dataset ($\approx 24K$ frames). In all, running time is more than two orders of magnitude faster than \cite{hyperlapse}.

\subsubsection{User Study}

We compare the results of \egosns, first and second order smoothness formulations, with na\"{\i}ve fast forward with $10\times$ speedup, implemented by sampling the input video uniformly. For \egos the speed is not directly controlled but is targeted for $10\times$ speedup by setting $K_{flow}$ to be $10$ times the average optical flow magnitude of the sequence.

We conducted a user study to compare our results with the baseline methods. We sampled short clips (5-10 seconds each) from the output of the three methods at hand. We made sure the clips start and end at the same geographic location. We showed each of the 35 subjects several pairs of clips, before stabilization, chosen at random. We asked the subjects to state which of the clips is better in terms of stability and continuity. The majority ($75\%$) of the subjects preferred the output of \egos with first-order shakiness term over the na\"{\i}ve baseline. On top of that, $68\%$ preferred the output of \egos using second-order shakiness term over the output using first-order shakiness term.

To evaluate the effect of video stabilization on the \egos output, we tested three commercial video stabilization tools: (i) Adobe Warp Stabilizer (ii) Deshaker \footnote[2]{http://www.guthspot.se/video/deshaker.htm} (iii) Youtube's Video stabilizer. We have found that  Youtube's stabilizer gives the best results on challenging fast forward videos \footnote[3]{We attribute this to the fact that Youtube's stabilizer does not depend upon long feature trajectories, which are scarce in sub-sampled video as ours.}. We stabilized the output clips using Youtube's stabilizer and asked our 35 subjects to repeat process described above. Again, the subjects favored the output of \egosns.

\subsubsection{Quantitative Evaluation}

We quantify the performance of \egos using the following measures. We measure the deviation of the output from the desired speedup. We found that measuring the speedup by taking the ratio between the number of input and output frames is misleading, because one of the features of \egos is to take large skips when the magnitude of the optical flow is rather low. We therefore measure the effective speedup as the median frame skip.

Additional measure is the reduction in epipole jitter between consecutive output frames (or FOE if F-Matrix cannot be estimated). We differentiate the locations of the epipole (temporally). The mean magnitude of the derivative gives us the amount of jitter between consecutive frames in the output. We measure the jitter for our method as well for naive $10\times$ uniform sampling and calculate the percentage improvement in jitter over competition.

Table \ref{tb:ff} shows the quantitative results for frame skip and epipole smoothness. There is a huge improvement in jitter by our algorithm. We note that the standard method to quantify video stabilization algorithms is to measure crop and distortion ratios. However since we jointly model fast forward and stabilization such measures are not applicable. The other method could have been to post process the output video with a standard video stabilization algorithm and measure these factors. Better measures might indicate better input to stabilization or better output from preceding sampling. However, most stabilization algorithms rely on trajectories and fail on resampled video with large view difference. The only successful algorithm was Youtube's stabilizer but it did not give us these measures.

\subsubsection{Limitations}

\begin{figure}[t]
    \centering
    \includegraphics[width=0.4\columnwidth]{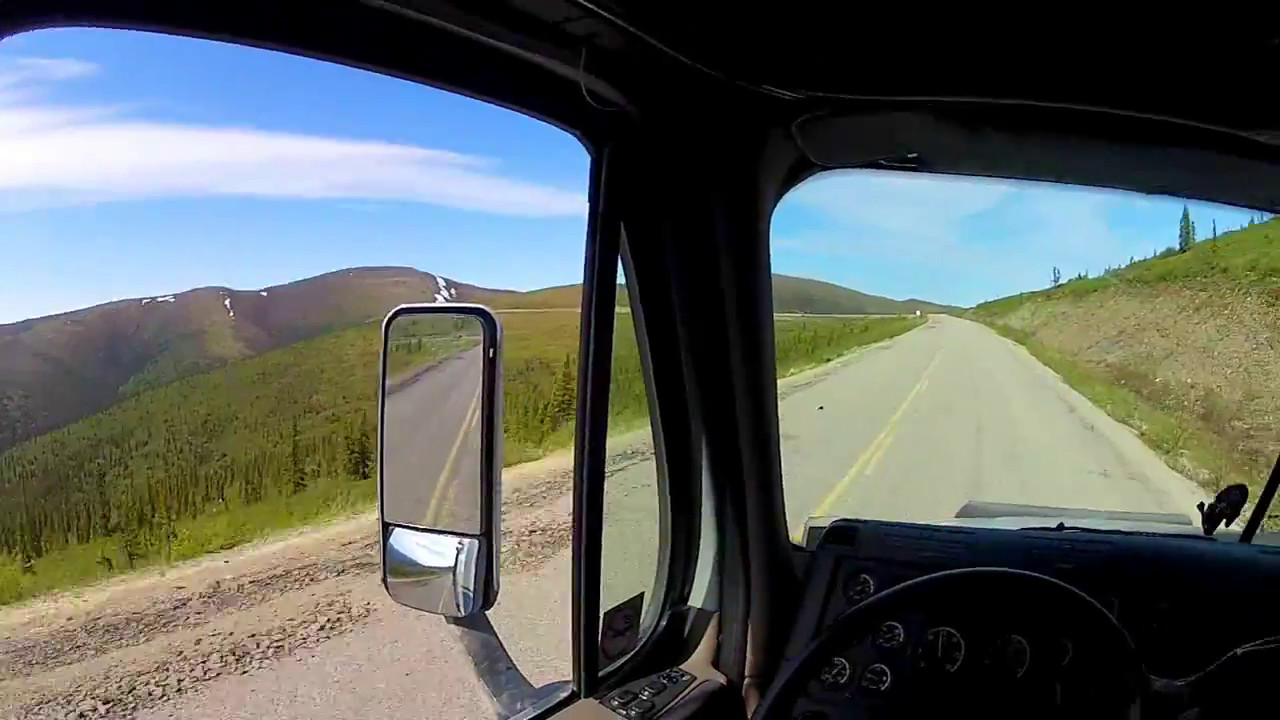}
    \includegraphics[width=0.4\columnwidth]{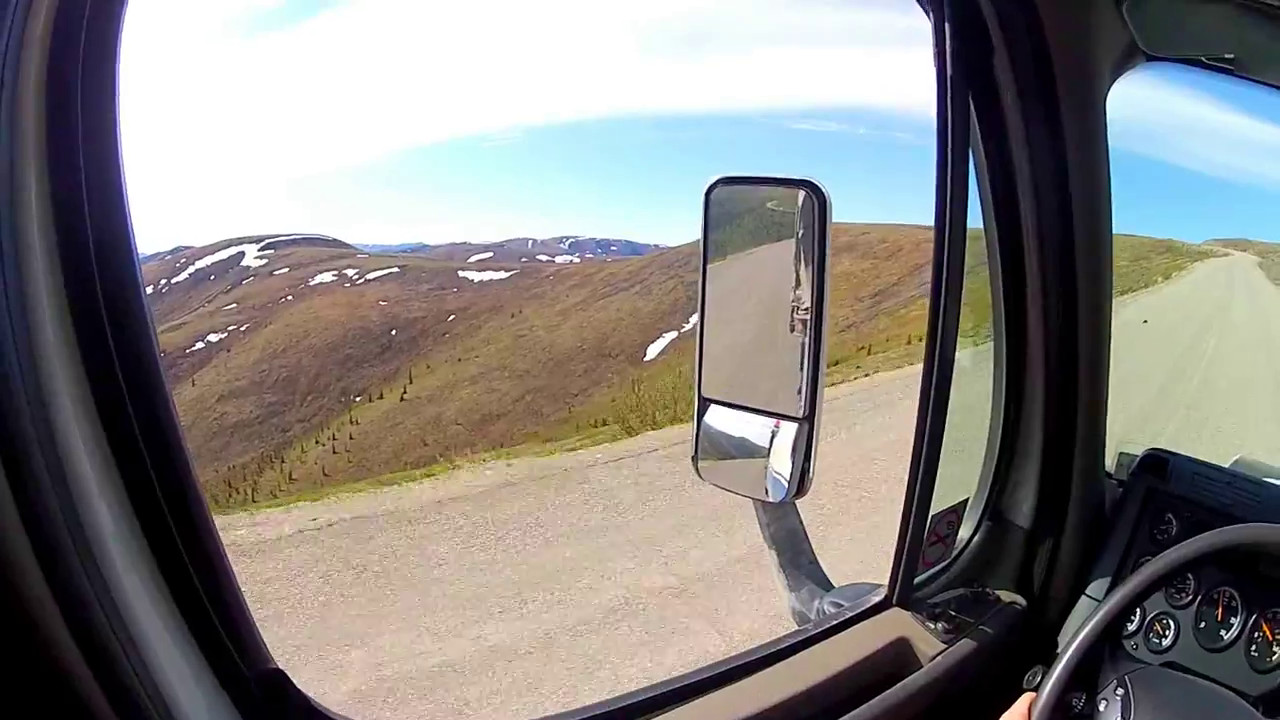}
    \caption{A failure case for the proposed method showing two sample frames from an input sequence. The frame to frame optical flow is mostly zero because of distant view and (relatively) static vehicle interior. However, since the driver shakes his head every few seconds, the average optical flow magnitude is high. The velocity term causes us to skip many frames until the desired $K_{flow}$ is met. Restricting the maximum frame skip by setting $\tau$ to a small value leads to arbitrary frames being chosen looking sideways, causing shake in the output video.}
    \label{fig:ff-failure-driving}
\end{figure}

One notable difference between \egos and traditional fast forward methods is that the number of output frames is not fixed. To adjust the effective speedup, the user can tune the velocity term by setting different values to $K_{flow}$. It should be noted, however, that not all speedup factors are possible without compromising the stability of the output. For example, consider a camera that toggles between looking straight and looking to the left every $10$ frames. Clearly, any speedup factor that is not a multiple of $10$ will introduce shake to the output. The algorithm chooses an optimal speedup factor which balances between the desired speedup and what can be achieved in practice on the specific input. Sequence `Driving' (Figure \ref{fig:ff-failure-driving}) presents an interesting failure case.

Another limitation of \egos is to handle long periods in which the camera wearer is static, hence, the camera is not translating. In these cases, both the fundamental matrix and the FOE estimations can become unstable, leading to wrong cost assignments (low penalty instead of high) to graph edges. The appearance and velocity terms are more robust and help reduce the number of outlier (shaky) frames in the output.

\subsection{\panohns}

In this section we show experiments to evaluate \panoh for single as well as multiple input videos. To evaluate the multiple videos case (Section \ref{sec:multiple_input_method}), we have used two types of video sets. The first type are videos sharing similar camera path on different times. We obtained the dataset of \cite{yedid_biometrics} suitable for this purpose. The second type are videos shot simultaneously by number of people wearing cameras and walking together. We scanned the dataset of \cite{ego_social} and found videos corresponding to a few minutes of a group walking together towards an amusement park. In addition, we choreographed two videos of this type by ourselves. We will release these videos upon paper acceptance. The videos were shot using a GoPro3+ camera. Table \ref{tb:ff_sequences} gives the resolution, FPS, length and source of the videos used in our experiments.

\renewcommand{\tabcolsep}{0.1cm}
\begin{table}[t]
    \caption{Comparing field of view (FOV): We measure cropping of output frame output by various methods. The percentages indicate the average area of the cropped image from the original input image, measured on 10 randomly sampled output frames from each sequence. The same frames were used for all the five methods. The naive, \egosns (ES), and \panohns (PH) outputs were stabilized using YouTube stabilizer \cite{youtube_stabilizer}. Real-time Hyperlapse \cite{rt-hyperlapse} output was created using the desktop version of the Hyperlapse Pro. app. The output of Hyperlapse \cite{hyperlapse} is only available for their dataset. We observe improvements in all the examples except `walking2', in which the camera is very steady.}
\label{tb:crop-size}
    \centering
    \begin{tabular}{lcccccc}
    \toprule[1.5pt]
        \specialcell{\bf \tabletext Name}  & \specialcell{\bf \tabletext Exp. No.}  &  \specialcell{\bf \tabletext Naive }   & \specialcell{\bf \tabletext \cite{rt-hyperlapse} } & \specialcell{\bf \tabletext \cite{hyperlapse} } & \specialcell{\bf \tabletext ES }  & \specialcell{\bf \tabletext PH} \\ \midrule

        \tabletext Bike3	&  \tabletext S1 &\specialcell{\tabletext 45\%}	&\specialcell{\tabletext 32\%}	&\specialcell{\tabletext 65\%}	&\specialcell{\tabletext 33\%} &\specialcell{\tabletext 99\%}		\\

		\tabletext Walking1	&  \tabletext S2 &\specialcell{\tabletext 52\%}	&\specialcell{\tabletext 68\%}	&\specialcell{\tabletext 68\%}	&\specialcell{\tabletext 40\%}	&\specialcell{\tabletext 95\%}		\\

		\tabletext Walking2	&  \tabletext S3 &\specialcell{\tabletext 67\%}	&\specialcell{\tabletext N/A}	&\specialcell{\tabletext N/A}	&\specialcell{\tabletext 43\%}	&\specialcell{\tabletext 66\%}	\\

		\tabletext Walking3	&  \tabletext S4 &\specialcell{\tabletext 71\%}	&\specialcell{\tabletext N/A}	&\specialcell{\tabletext N/A}	&\specialcell{\tabletext 54\%}	&\specialcell{\tabletext 102\%}	\\

		\tabletext Walking4	&  \tabletext S5 &\specialcell{\tabletext 68\%}	&\specialcell{\tabletext N/A}	&\specialcell{\tabletext N/A}	&\specialcell{\tabletext 44\%}	&\specialcell{\tabletext 109\%}	\\

		\tabletext Running &  \tabletext S6 	&\specialcell{\tabletext 50\%}	&\specialcell{\tabletext 75\%}	&\specialcell{\tabletext N/A}	&\specialcell{\tabletext 43\%}	&\specialcell{\tabletext 101\%}	\\

\bottomrule[1.5pt] \\
    \end{tabular}
\end{table}

\begin{figure*}[t]
\begin{tabular}{cccc}
\includegraphics[height=2.8cm,keepaspectratio]{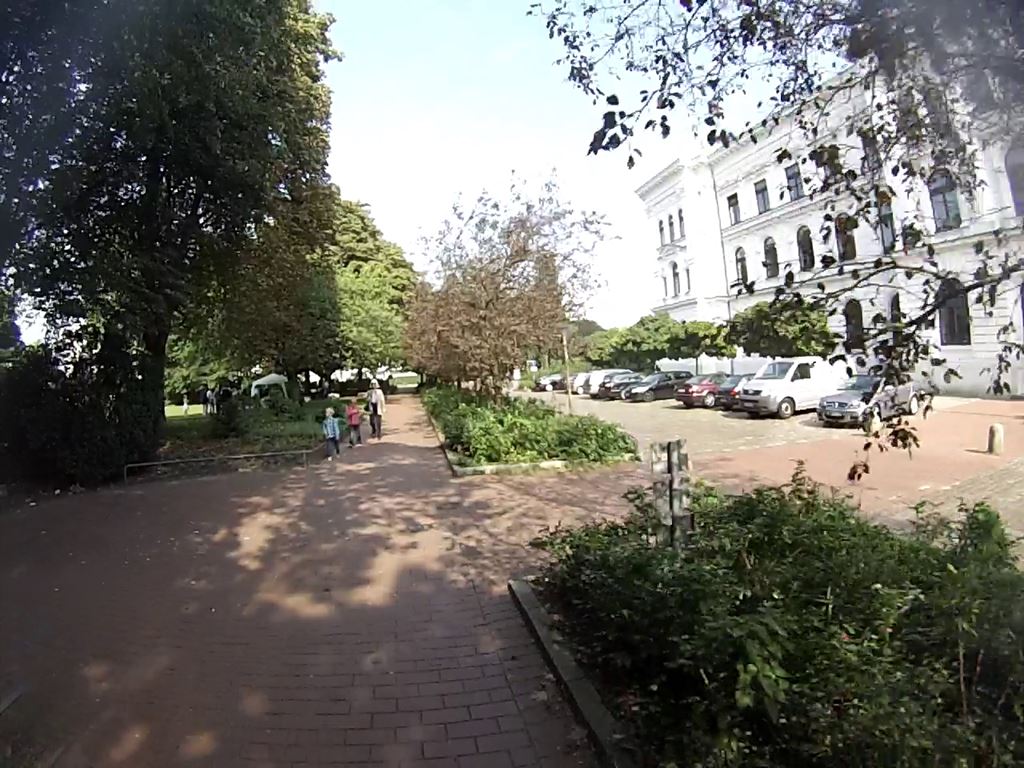}
& \includegraphics[height=2.8cm,keepaspectratio]{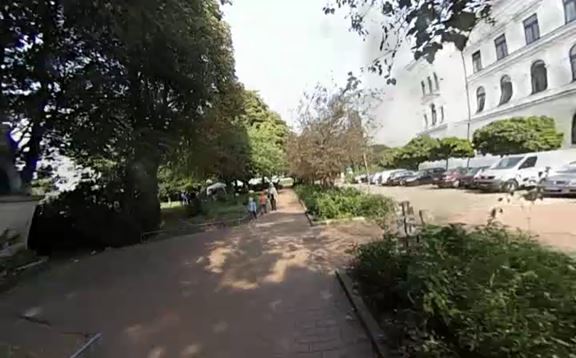}
& \includegraphics[height=2.8cm,keepaspectratio]{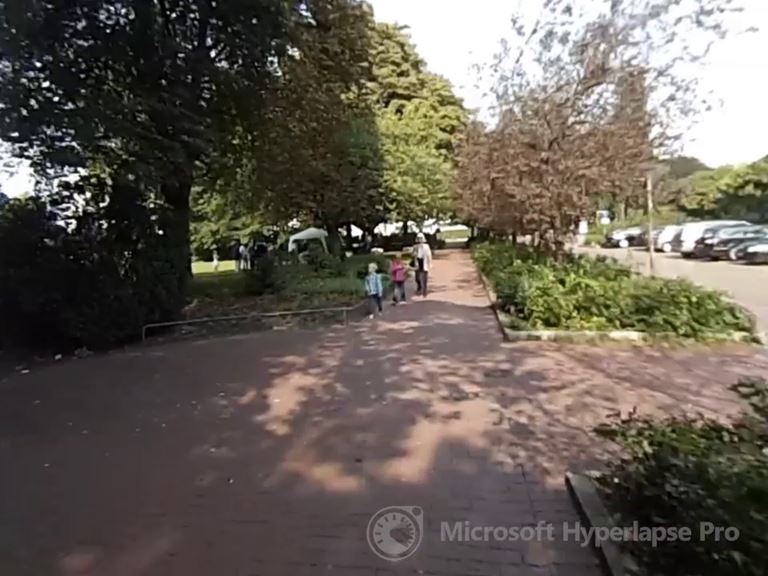}
& \includegraphics[height=2.8cm,keepaspectratio]{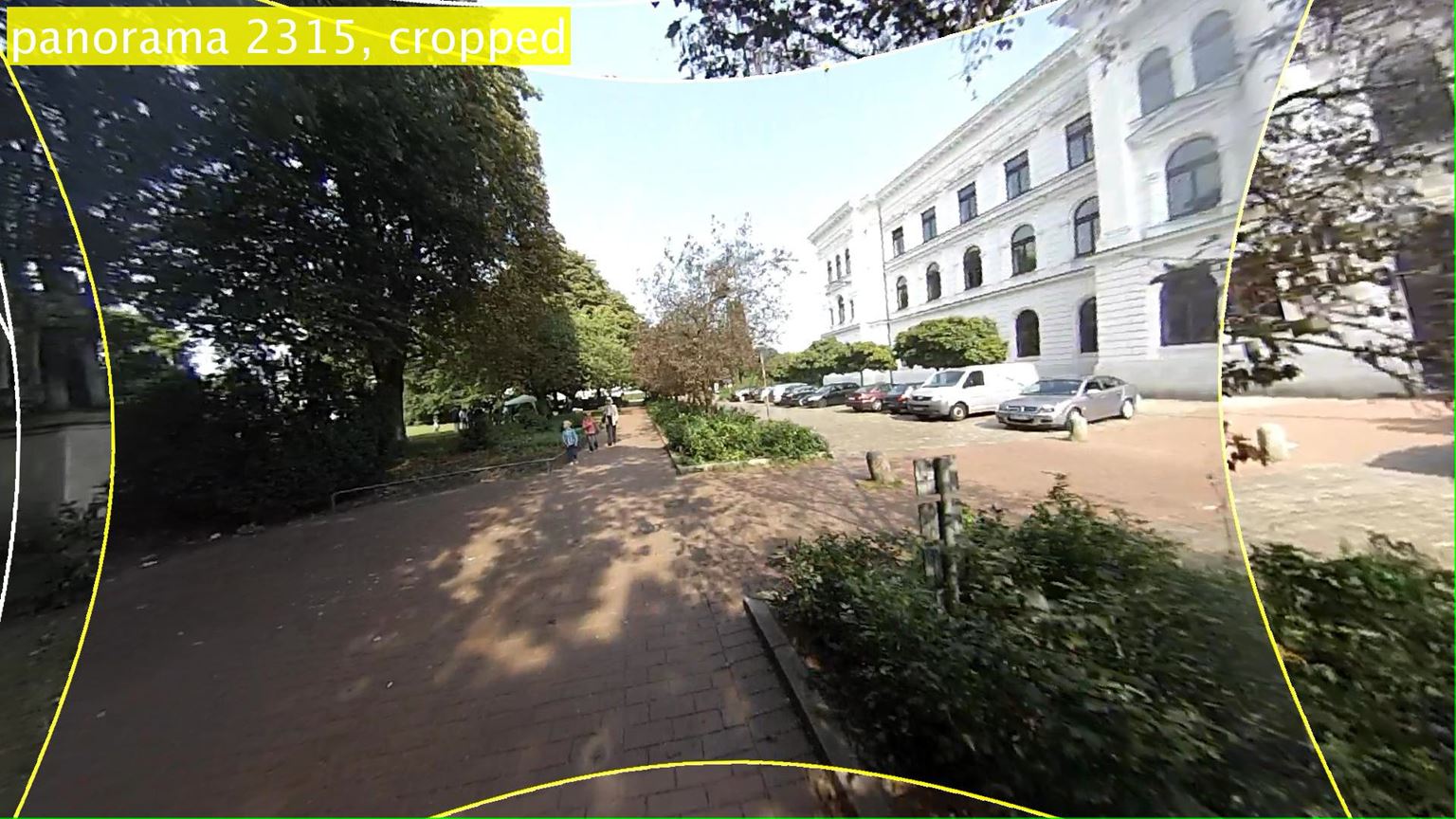} \\

(a) & (b) & (c) & (d)
\end{tabular}

\caption{Comparing FOV of hyperlapse frames, corresponding to approximately same input frames from sequence `Bike1'. For best viewing zoom to 800\%. \underline{Columns:} (a) Original frame and output of \egosns. (b) Output of \cite{hyperlapse}. Cropping and rendering errors are clearly visible. (c) Output of \cite{rt-hyperlapse} suffering from strong cropping. (d) Output of our method, having the largest FOV.}
\label{fig:frame_comparison}
\end{figure*}

\subsection{Implementation Details}

We have implemented \panoh in Matlab and run it on a single PC with no GPU support. For tracking we use Matlab's built in SURF feature points detector and tracker. We found the homography between frames using RANSAC. This is a time consuming step since it requires calculating transformations from every frame which is a candidate for a panorama center, to every other frame in the temporal window around it (typically $\omega=50$). In addition, we find homographies to other frames that may serve as other panorama centers (before/after the current frame), in order to calculate the Shakiness cost of a transition between them. We avoid creating the actual panoramas after the sampling step to reduce runtime. However, we still have to calculate the panorama's $FOV$ as it is part of our cost function. We resolved to created a mask of the panorama, which is faster than creating the panorama itself. The parameters of the cost function in Eq.~(\ref{eq:cost}) were set to $\alpha=1\cdot10^{7}$, $\beta=5\cdot10^{6}$, $\gamma=1$ and $\lambda=15$ for the crop window smoothness. Our $cross-video$ term was multiplied by the constant $2$. We used those parameters both for the single and multi video scenarios. The input and output videos are given at the project's website.

\subsection{Runtime}

The following runtimes were measured with the setup described in the previous section on a 640$\times$480 resolution video, processing a single input video. Finding the central images and calculating the Shakiness cost takes 200ms per frame, each.
Calculating the FOV term takes 100ms per frame on average. Finding the shortest path takes a few seconds for the entire sequence.
Sampling and panorama creation takes 3 seconds per panorama, and the total time depends on the speed up from the original video i.e. the ratio between number of panoramas and length of the input. For a typical $\times10$ speed this amounts to ~300ms. The total runtime is 1.5-2 seconds per frame with an unoptimized Matlab implementation. In the multi-input video cases the runtime grows linearly with the number of input sequences.


\renewcommand{\tabcolsep}{0.3cm}
\begin{table}[t]
    \caption{Evaluation of the contribution of multiple videos to the FOV. The crop size was measured twice: once with the single video algorithm, with the video in the first column as input, and once with the multi video algorithm.
    }
	\label{tb:multi-video}
    \centering
    \begin{tabular}{lcccccc}
    \toprule[1.5pt]
        \specialcell{\bf \tabletext }  & \specialcell{\bf \tabletext }  &  \specialcell{\bf \tabletext Ours }   & \specialcell{\bf \tabletext Number } & \specialcell{\bf \tabletext Ours  } \\

		\specialcell{\bf \tabletext Name}  & \specialcell{\bf \tabletext Exp. No.}  &  \specialcell{\bf \tabletext Single}   & \specialcell{\bf \tabletext of Videos  } & \specialcell{\bf \tabletext  Multi }  \\ \midrule

        \tabletext Walking2	&  \tabletext M1 &\specialcell{\tabletext 67\%}	&\specialcell{\tabletext 4}	&\specialcell{\tabletext 140\%}	\\

        \tabletext Walking5	&  \tabletext M2 &\specialcell{\tabletext 90\%}	&\specialcell{\tabletext 2}	&\specialcell{\tabletext 98\%}	\\

        \tabletext Walking7	&  \tabletext M3 &\specialcell{\tabletext 107\%}	&\specialcell{\tabletext 2}	&\specialcell{\tabletext 118\%}	\\
\bottomrule[1.5pt] \\
    \end{tabular}
\end{table}

\begin{figure}[t]
\centering
\subfigure{\includegraphics[height=2.35cm]{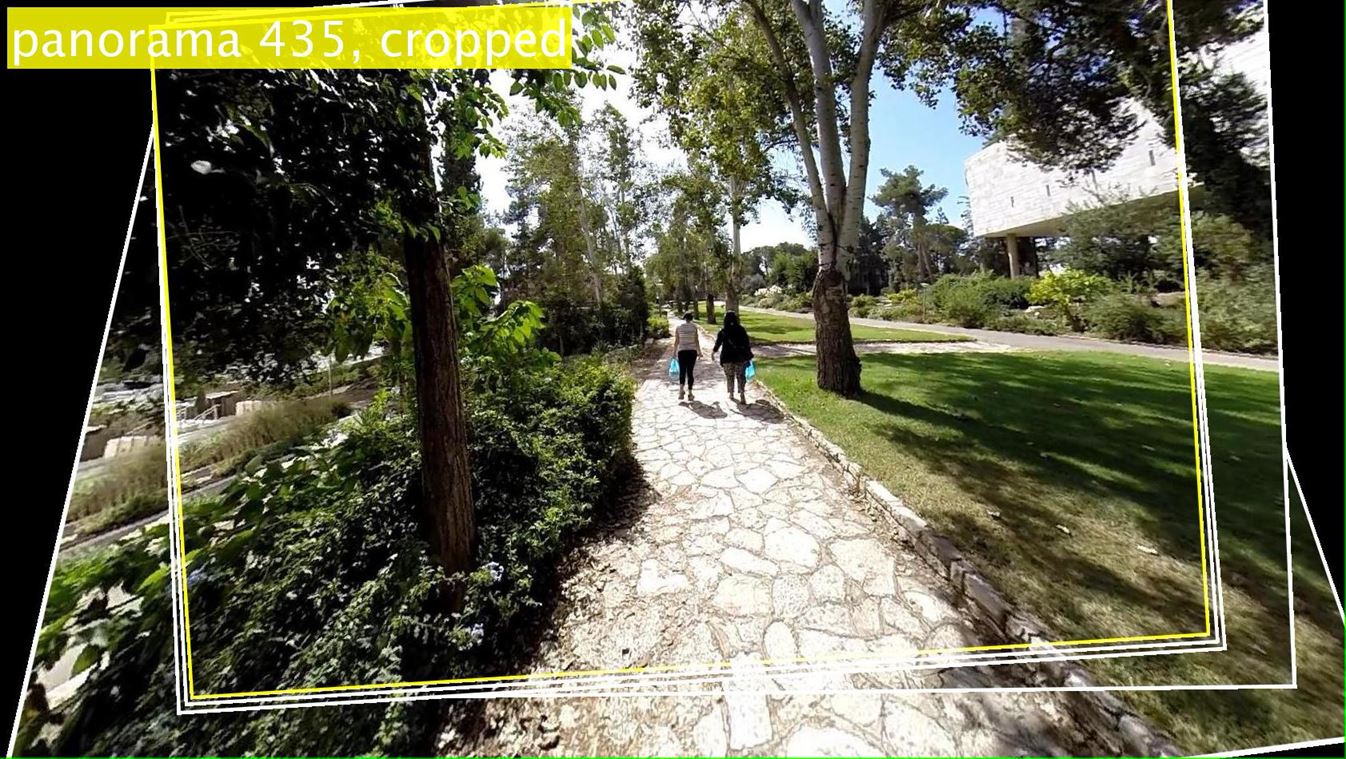} }
\subfigure{\includegraphics[height=2.35cm]{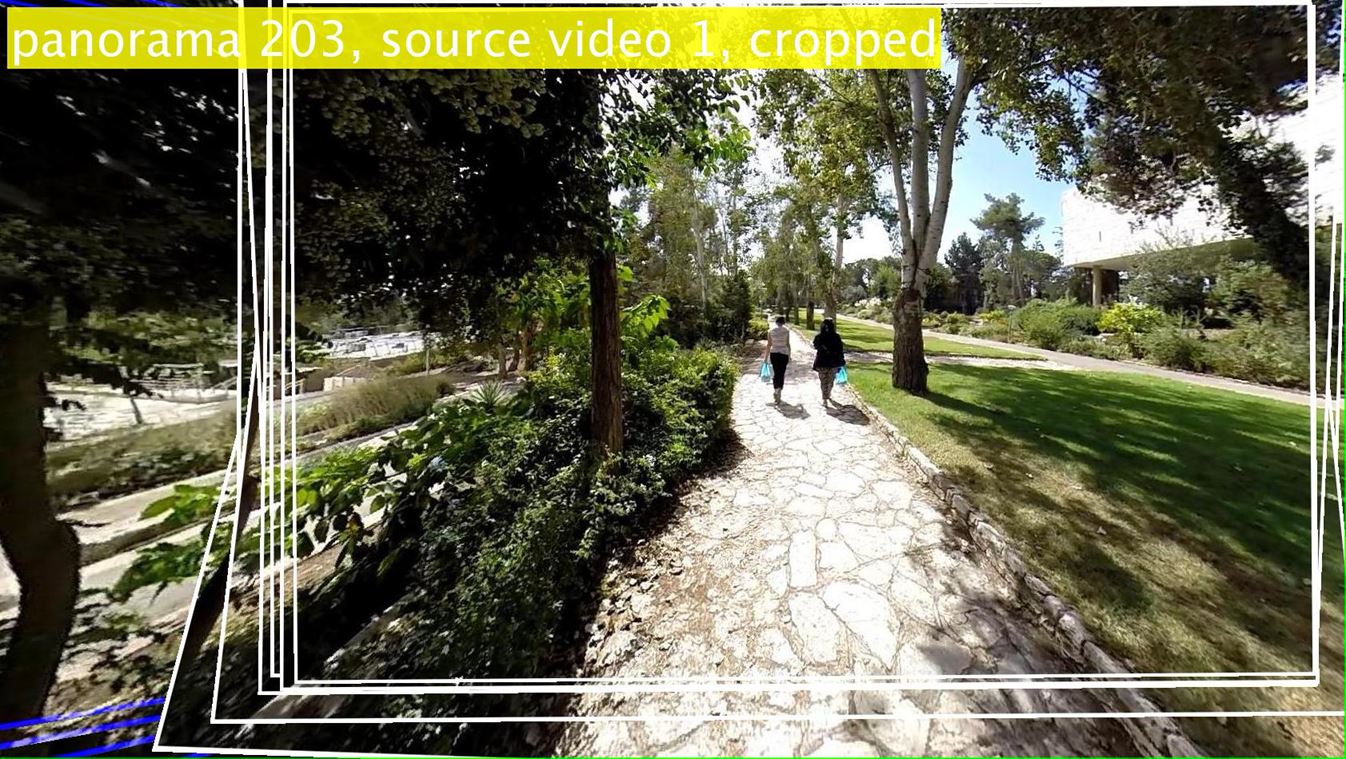} }
    \caption{Comparing field-of-view of panoramas generated from single (left) and multi (right) video \panohns. Multi video \panoh is able to successfully collate content from different videos for enhanced field of view.}
    \label{fig:single-multi-crop-comparison}
\end{figure}

\begin{figure*}[t]
	\centering
\subfigure{\includegraphics[height=2.8cm]{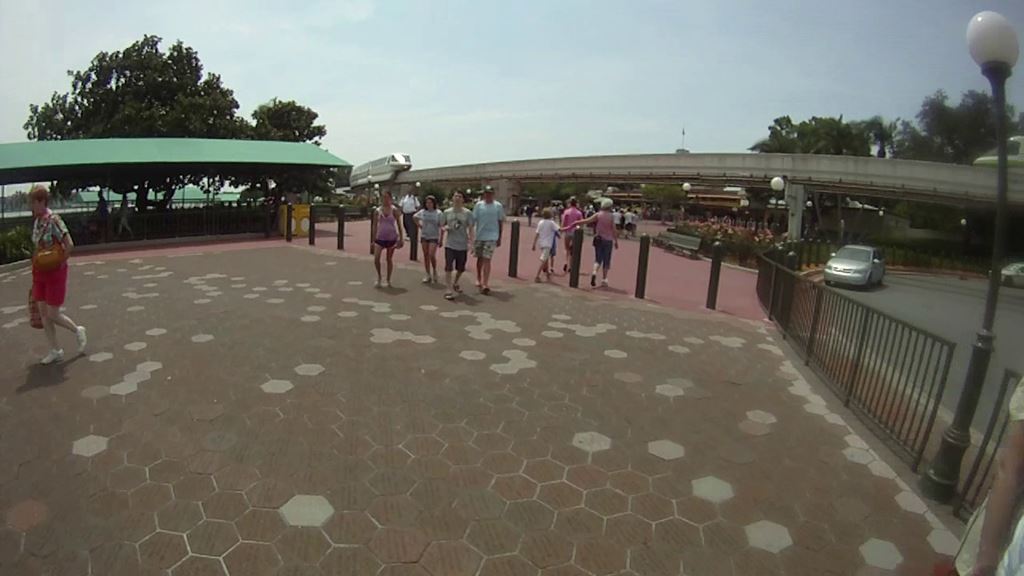}} \subfigure{\includegraphics[height=2.8cm]{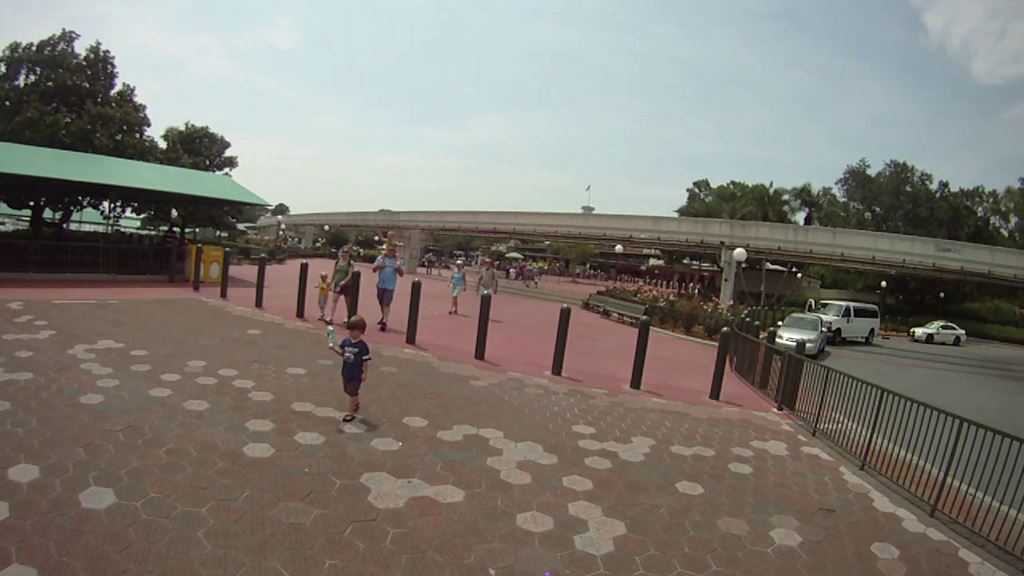}}
\subfigure{\includegraphics[height=2.8cm]{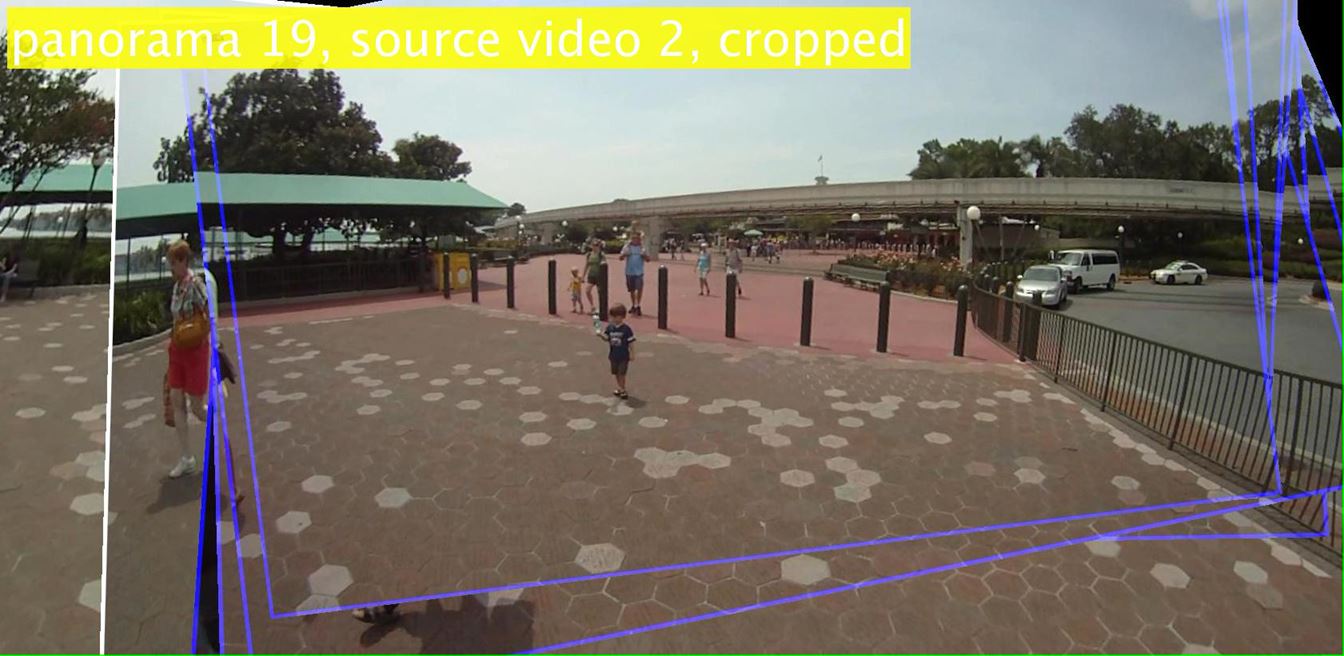}}

    \caption{\panohns: Left and middle are two input spatially neighboring frames from different videos. Right is the output frame generated by \panohns. The blue lines indicate frames coming from the same video as the middle frame (Walking6), while the white lines indicate frames from the other video (Walking5). Notice that while a lady can be observed in one and a child in another, both are visible in the output frames. The stitching errors are due to misalignment of the frames. We did not have the camera information for these sequences and could not perform lens distortion correction}
    \label{fig:single-multi}
\end{figure*}

\subsection{Evaluation}
\label{sec:evaluation}

The main contribution of \panoh to the hyperlapse community is the increased field of view (FOV) over  existing methods. To evaluate it we measure the output resolution (i.e. the crop size) of the baseline hyperlapse methods on the same sequence. The crop is a side-effect of stabilization: without crop, stabilization introduces ``empty" pixels to the field of view. The cropping ensures to limit the output frame to the intersection of several FOVs, which can be substantially smaller than the FOV of each frame depending on the shakiness of the video.

The crop size is not constant throughout the whole output video, hence it should be compared individually between output frames. Because of the frame sampling, an output frame with one method is not guaranteed to appear in the output of another method. Therefore, we randomly sampled frames for each sequence until we had 10 frames that appear in all output methods. For a panorama we considered its central frame. We note that the output of \cite{hyperlapse} is rendered from several input frames, and does not have any dominant frame. We therefore tried to pick frames corresponding to the same geographical location in the other sequences. Our results are summarized in Tables \ref{tb:crop-size} and \ref{tb:multi-video}. It is clear that in terms of FOV we outperform most of the baseline methods on most of the sequences. The contribution of multiple videos to the FOV is illustrated in Figure \ref{fig:single-multi-crop-comparison}.

The naive fast forward, \egosns, and \panoh outputs were stabilized using YouTube stabilizer. Real-time Hyperlapse \cite{rt-hyperlapse} output was created using the desktop version of the Hyperlapse Pro. app. The output of Hyperlapse \cite{hyperlapse} is only available for their dataset.

\paragraph{Failure case} On sequence Walking2 the naive results get the same crop size as our method (see Table \ref{tb:crop-size}). We attribute this to the exceptionally steady forward motion of the camera, almost as if it is not mounted on the photographer head while walking. Obviously, without the shake \panoh can not extend the field of view significantly.

\subsection{\panoh from Multiple Videos}

Fig.~\ref{fig:single-multi-crop-comparison} shows a sample frame from the output generated by our algorithm using sequences `Walking 7' and `Walking 8'. Comparison with panoramic hyperlapse generated from single video clearly shows that our method is able to assemble content from frames from multiple videos for enhanced field of view. We quantify the improvement in FOV using the crop ratio of the output video on various publicly and self shot test sequences. Table~\ref{tb:multi-video} gives the detailed comparison.

Multi Video \panoh can also be used to summarize contents from multiple videos. Fig.~\ref{fig:single-multi} shows an example panorama generated from sequences `Walking 5' and `Walking 6' from the dataset released by \cite{ego_social}. While a lady is visible in one video and a child in another, both persons appear in the output frame at the same time.

When using multiple videos, each panorama in the \panoh is generated from many frames, as much as 150 frames if we use three videos and a temporal window of 50 frames. With this wealth of frames, we can filter out some frames with undesired properties. For example, if privacy is a concern, we can remove from the panorama all frames having a recognizable face or a readable license plate.

\section{Conclusion}
\label{sec:conclusion}

We propose a novel frame sampling technique to produce stable fast forward egocentric videos. Instead of the demanding task of $3D$ reconstruction and rendering used by the best existing methods, we rely on simple computation of the epipole or the FOE. The proposed framework is very efficient, which makes it practical for long egocentric videos. Because of its reliance on simple optical flow, the method can potentially handle difficult egocentric videos, where methods requiring 3D reconstruction may not be reliable.

We also present \panohns, a method to create hyperlapse videos having a large field-of-view. While in \egos we drop unselected (outlier) frames, in \panohns, we use them to increase the field of view in the output video. In addition, \panoh naturally supports the processing of multiple videos together, extending the output field of view even further, as well as allowing to consume multiple such videos in less time. The large number of frames used for each panorama also allows to remove undesired objects from the output.


\bibliographystyle{IEEEtran}
\bibliography{IEEEabrv,egosampling}

\begin{thebibliography}{10}
\providecommand{\url}[1]{#1}
\csname url@samestyle\endcsname
\providecommand{\newblock}{\relax}
\providecommand{\bibinfo}[2]{#2}
\providecommand{\BIBentrySTDinterwordspacing}{\spaceskip=0pt\relax}
\providecommand{\BIBentryALTinterwordstretchfactor}{4}
\providecommand{\BIBentryALTinterwordspacing}{\spaceskip=\fontdimen2\font plus
\BIBentryALTinterwordstretchfactor\fontdimen3\font minus
  \fontdimen4\font\relax}
\providecommand{\BIBforeignlanguage}[2]{{%
\expandafter\ifx\csname l@#1\endcsname\relax
\typeout{** WARNING: IEEEtran.bst: No hyphenation pattern has been}%
\typeout{** loaded for the language `#1'. Using the pattern for}%
\typeout{** the default language instead.}%
\else
\language=\csname l@#1\endcsname
\fi
#2}}
\providecommand{\BIBdecl}{\relax}
\BIBdecl

\bibitem{grauman-story}
Z.~Lu and K.~Grauman, ``Story-driven summarization for egocentric video,'' in
  \emph{CVPR}, 2013.

\bibitem{grauman-important-people}
Y.~J. Lee, J.~Ghosh, and K.~Grauman, ``Discovering important people and objects
  for egocentric video summarization,'' in \emph{CVPR}, 2012.

\bibitem{rehg_gaze}
J.~Xu, L.~Mukherjee, Y.~Li, J.~Warner, J.~M. Rehg, and V.~Singh, ``Gaze-enabled
  egocentric video summarization via constrained submodular maximization,'' in
  \emph{CVPR}, 2015.

\bibitem{us_egoseg}
Y.~Poleg, C.~Arora, and S.~Peleg, ``Temporal segmentation of egocentric
  videos,'' in \emph{CVPR}, 2014, pp. 2537--2544.

\bibitem{compact_cnn}
\BIBentryALTinterwordspacing
Y.~Poleg, A.~Ephrat, S.~Peleg, and C.~Arora, ``Compact {CNN} for indexing
  egocentric videos,'' in \emph{WACV}, 2016. [Online]. Available:
  \url{http://arxiv.org/abs/1504.07469}
\BIBentrySTDinterwordspacing

\bibitem{kitani}
K.~M. Kitani, T.~Okabe, Y.~Sato, and A.~Sugimoto, ``Fast unsupervised
  ego-action learning for first-person sports videos,'' in \emph{CVPR}, 2011.

\bibitem{ryoo_pooled}
M.~S. Ryoo, B.~Rothrock, and L.~Matthies, ``Pooled motion features for
  first-person videos,'' in \emph{CVPR}, 2015, pp. 896--904.

\bibitem{hyperlapse}
\BIBentryALTinterwordspacing
J.~Kopf, M.~Cohen, and R.~Szeliski, ``First-person hyperlapse videos,'' in
  \emph{SIGGRAPH}, vol.~33, no.~4, August 2014. [Online]. Available:
  \url{http://research.microsoft.com/apps/pubs/default.aspx?id=230645}
\BIBentrySTDinterwordspacing

\bibitem{Petrovic:2005}
N.~Petrovic, N.~Jojic, and T.~S. Huang, ``Adaptive video fast forward,''
  \emph{Multimedia Tools Appl.}, vol.~26, no.~3, pp. 327--344, Aug. 2005.

\bibitem{rt-hyperlapse}
N.~Joshi, W.~Kienzle, M.~Toelle, M.~Uyttendaele, and M.~F. Cohen, ``Real-time
  hyperlapse creation via optimal frame selection,'' in \emph{SIGGRAPH},
  vol.~34, no.~4, 2015, p.~63.

\bibitem{egosampling}
Y.~Poleg, T.~Halperin, C.~Arora, and S.~Peleg, ``Egosampling: Fast-forward and
  stereo for egocentric videos,'' in \emph{CVPR}, 2015, pp. 4768--4776.

\bibitem{hyperlapse-dataset}
\BIBentryALTinterwordspacing
J.~Kopf, M.~Cohen, and R.~Szeliski, ``{First-person Hyperlapse Videos -
  Supplemental Material}.'' [Online]. Available:
  \url{http://research.microsoft.com/en-us/um/redmond/projects/hyperlapse/supplementary/index.html}
\BIBentrySTDinterwordspacing

\bibitem{grauman-snap-points}
B.~Xiong and K.~Grauman, ``Detecting snap points in egocentric video with a web
  photo prior,'' in \emph{ECCV}, 2014.

\bibitem{content_preserving_warps}
F.~Liu, M.~Gleicher, H.~Jin, and A.~Agarwala, ``Content-preserving warps for 3d
  video stabilization,'' in \emph{SIGGRAPH}, 2009.

\bibitem{vid3d_stab_depth}
S.~Liu, Y.~Wang, L.~Yuan, J.~Bu, P.~Tan, and J.~Sun, ``Video stabilization with
  a depth camera,'' in \emph{CVPR}, 2012.

\bibitem{youtube_stabilizer}
M.~Grundmann, V.~Kwatra, and I.~Essa, ``Auto-directed video stabilization with
  robust l1 optimal camera paths,'' in \emph{CVPR}, 2011.

\bibitem{subspace_vid_stab}
F.~Liu, M.~Gleicher, J.~Wang, H.~Jin, and A.~Agarwala, ``Subspace video
  stabilization,'' in \emph{SIGGRAPH}, 2011.

\bibitem{BundledPaths2013}
S.~Liu, L.~Yuan, P.~Tan, and J.~Sun, ``Bundled camera paths for video
  stabilization,'' in \emph{SIGGRAPH}, 2013.

\bibitem{steadyflow}
------, ``Steadyflow: Spatially smooth optical flow for video stabilization,''
  2014.

\bibitem{raanan}
A.~Goldstein and R.~Fattal, ``Video stabilization using epipolar geometry,'' in
  \emph{SIGGRAPH}, 2012.

\bibitem{eyalofek_stab}
Y.~Matsushita, E.~Ofek, W.~Ge, X.~Tang, and H.~Shum, ``Full-frame video
  stabilization with motion inpainting,'' \emph{{IEEE} Trans. PAMI}, vol.~28,
  no.~7, pp. 1150--1163, 2006.

\bibitem{jiang2015video_stitching}
W.~Jiang and J.~Gu, ``Video stitching with spatial-temporal content-preserving
  warping,'' in \emph{CVPR Workshops}, 2015, pp. 42--48.

\bibitem{yedid_curation}
Y.~Hoshen, G.~Ben{-}Artzi, and S.~Peleg, ``Wisdom of the crowd in egocentric
  video curation,'' in \emph{CVPR Workshops}, 2014, pp. 587--593.

\bibitem{arev_auto_social_cams}
I.~Arev, H.~S. Park, Y.~Sheikh, J.~K. Hodgins, and A.~Shamir, ``Automatic
  editing of footage from multiple social cameras,'' 2014.

\bibitem{hartley_book}
R.~Hartley and A.~Zisserman, \emph{Multiple View Geometry in Computer Vision},
  2nd~ed.\hskip 1em plus 0.5em minus 0.4em\relax Cambridge University Press,
  2003.

\bibitem{lsdslam}
J.~Engel, T.~Schöps, and D.~Cremer, ``{LSD-SLAM}: Large-scale direct monocular
  {SLAM},'' in \emph{ECCV}, 2014.

\bibitem{svo}
C.~Forster, M.~Pizzoli, and D.~Scaramuzza, ``Svo: Fast semi-direct monocular
  visual odometry,'' in \emph{ICRA}, 2014.

\bibitem{technion-foe}
D.~Sazbon, H.~Rotstein, and E.~Rivlin, ``Finding the focus of expansion and
  estimating range using optical flow images and a matched filter.''
  \emph{Machine Vision Applications}, vol.~15, no.~4, pp. 229--236, 2004.

\bibitem{emd}
O.~Pele and M.~Werman, ``Fast and robust earth mover's distances,'' in
  \emph{ICCV}, 2009.

\bibitem{dijkstra}
E.~Dijkstra, ``A note on two problems in connexion with graphs,''
  \emph{NUMERISCHE MATHEMATIK}, vol.~1, no.~1, 1959.

\bibitem{szeliski2006image}
R.~Szeliski, ``Image alignment and stitching: A tutorial,'' \emph{Foundations
  and Trends in Computer Graphics and Vision}, vol.~2, no.~1, pp. 1--104, 2006.

\bibitem{zelnik2007automating}
L.~Zelnik-Manor and P.~Perona, ``Automating joiners,'' in \emph{Proceedings of
  the 5th international symposium on Non-photorealistic animation and
  rendering}.\hskip 1em plus 0.5em minus 0.4em\relax ACM, 2007, pp. 121--131.

\bibitem{ocamlib}
D.~Scaramuzza, A.~Martinelli, and R.~Siegwart, ``A toolbox for easily
  calibrating omnidirectional cameras,'' in \emph{Intelligent Robots and
  Systems, 2006 IEEE/RSJ International Conference on}, Oct 2006, pp.
  5695--5701.

\bibitem{youtube-running1}
\BIBentryALTinterwordspacing
``{Ayala Triangle Run with GoPro Hero 3+ Black Edition}.'' [Online]. Available:
  \url{https://www.youtube.com/watch?v=WbWnWojOtIs}
\BIBentrySTDinterwordspacing

\bibitem{youtube-driving2}
\BIBentryALTinterwordspacing
``{GoPro Trucking! - Yukon to Alaska 1080p}.'' [Online]. Available:
  \url{https://www.youtube.com/watch?v=3dOrN6-V7V0}
\BIBentrySTDinterwordspacing

\bibitem{ego_social}
A.~Fathi, J.~K. Hodgins, and J.~M. Rehg, ``Social interactions: A first-person
  perspective,'' in \emph{CVPR}, 2012.

\bibitem{lk}
B.~D. Lucas and T.~Kanade, ``An iterative image registration technique with an
  application to stereo vision,'' in \emph{IJCAI}, vol.~2, 1981.

\bibitem{visualsfm}
``{VisualSFM : A Visual Structure from Motion System, Changchang Wu,
  http://ccwu.me/vsfm/}.''

\bibitem{yedid_biometrics}
\BIBentryALTinterwordspacing
Y.~Hoshen and S.~Peleg, ``An egocentric look at video photographer identity,''
  in \emph{CVPR}, 2016. [Online]. Available:
  \url{http://arxiv.org/abs/1411.7591}
\BIBentrySTDinterwordspacing

\end{thebibliography}

\end{document}